\documentclass{article} % For LaTeX2e
\usepackage{iclr2027_conference,times}

\usepackage{amsmath,amsfonts,bm}

\def\eqref#1{equation~\ref{#1}}
\def\Eqref#1{Equation~\ref{#1}}
\def\1{\bm{1}}

\def\vmu{{\bm{\mu}}}

\def\va{{\bm{a}}}

\def\vc{{\bm{c}}}

\def\vp{{\bm{p}}}

\def\vz{{\bm{z}}}

\DeclareMathAlphabet{\mathsfit}{\encodingdefault}{\sfdefault}{m}{sl}
\SetMathAlphabet{\mathsfit}{bold}{\encodingdefault}{\sfdefault}{bx}{n}

\def\gC{{\mathcal{C}}}
\def\gD{{\mathcal{D}}}

\def\gZ{{\mathcal{Z}}}

\newcommand{\E}{\mathbb{E}}
\newcommand{\Ls}{\mathcal{L}}
\newcommand{\R}{\mathbb{R}}

\usepackage[utf8]{inputenc} % allow utf-8 input
\usepackage[T1]{fontenc}    % use 8-bit T1 fonts
\usepackage{hyperref}       % hyperlinks
\usepackage{url}            % simple URL typesetting
\usepackage{booktabs}       % professional-quality tables
\usepackage{amsfonts}       % blackboard math symbols
\usepackage{nicefrac}       % compact symbols for 1/2, etc.
\usepackage{microtype}      % microtypography
\usepackage{xcolor}         % colors
\usepackage{amsmath}
\usepackage{multirow}
\usepackage{graphicx}
\usepackage{pifont}
\usepackage{tcolorbox}      % for \tcbox
\usepackage{fontawesome5}   % for \faGlobe and \faGithub
\usepackage{subcaption}
\definecolor{codebg}{HTML}{F5F5F5}
\definecolor{codecolor}{HTML}{333333}

\title{Does Latent Planning Survive Point Clouds? Action-Conditioned JEPA World Models for Geometric Observations and Goals}

\author{%
  Fabio F. Oberweger\thanks{Joint first authors: F.F.O. led algorithm development, M.S. led data generation and environment design.}\hphantom{$^\ast$}\thanks{Correspondence to \texttt{fabio.oberweger@ait.ac.at}.}\hphantom{$^\dagger$}, Michael Schwingshackl$^\ast$ \& Markus Murschitz \\
  Center for Vision, Automation \& Control \\
  AIT Austrian Institute of Technology \\
}

\newcommand{\plewm}{Point-LeWM}
\newcommand{\pdjepa}{Point-Delta-JEPA}
\newcommand{\utoniawm}{Utonia-WM}
\newcommand{\voxwm}{Vox-WM}

\iclrfinalcopy % Uncomment for camera-ready version, but NOT for submission.
\begin{document}

\maketitle
\lhead{Preprint.}

% !!!!!!!!!!!!!!!!!!!!!!!!!!!!!!!!!!!!!!!!!!!!!!!!!!!!!!!!!!!!!!!!!!!!!!!!
% DON'T FORGET TO REMOVE ACKNOWLEDGEMENT FOR VSC
% REMOVE BEFORE WORKSHOP SUBMISSION 
\vspace{-2.5em}
\begin{center}
\href{https://fafraob.github.io/point-lewm/}{%
  \tcbox[
    on line,
    colback=codebg,
    colframe=codebg,
    coltext=codecolor,
    boxrule=0.4pt,
    arc=4pt,
    boxsep=2pt,
    left=4pt, right=4pt, top=2pt, bottom=2pt
  ]{\fontfamily{fvm}\selectfont\small\faGlobe\enspace Website}}
\href{https://github.com/fafraob/point-lewm}{%
  \tcbox[
    on line,
    colback=codebg,
    colframe=codebg,
    coltext=codecolor,
    boxrule=0.4pt,
    arc=4pt,
    boxsep=2pt,
    left=4pt, right=4pt, top=2pt, bottom=2pt
  ]{\fontfamily{fvm}\selectfont\small\faGithub\enspace Code}}
\end{center}
\vspace{2pt}

\begin{abstract}
Latent action world models let agents plan new behaviors at test time by predicting how actions change the environment, and joint-embedding predictive architectures (JEPAs) do so by forecasting future latent states rather than pixels. Yet nearly all such models see the world through a camera, even though robotic manipulation is fundamentally geometric: in robotics, goals for manipulation are traditionally specified by target object poses, not by images of the object once placed. We ask whether latent planning survives a shift from appearance to geometry, on the observation side as well as on the goal specifications side. To answer this, we extend the stable-worldmodel evaluation platform with simulated LiDAR-style raycast point clouds as a new sensor modality, and adapt three JEPA designs to point clouds: a frozen-encoder model built on Utonia features, a distribution-prior model based on LeWM, and an action-sensitive model based on Delta-JEPA. We further introduce a goal-encoding mechanism that constructs the goal latent from the current latent and a 3D target pose, removing the need for goal images or goal point clouds. A comparative evaluation of the different anti-collapse mechanisms shows that point-cloud world models can match their image-based counterparts, demonstrating that the modality shift from appearance to geometry is achievable. All models are released as open weights with open-source training and inference code, to make world-model planning accessible for LiDAR-driven and pose-directed robotic tasks.
\end{abstract}

\section{Introduction}
\label{sec:intro}

% A robot that can predict the consequences of its own actions need not be programmed for
% each task: control reduces to search. World models promise this for physical AI
% \citep{ha2018world,hafner2019learning,lecun2022path}: learn dynamics from raw experience,
% offline and reward-free, and plan new behaviors at test time. 
World models learn to predict how actions change the environment, allowing agents to plan new behaviors at test time instead of being trained for each task \citep{ha2018world,hafner2019learning,lecun2022path}.
Joint-embedding predictive
architectures (JEPAs) made this practical by predicting future \emph{latent} states
conditioned on actions rather than generating observations, whether over frozen features
(DINO-WM \citep{zhou2025dinowm}), distribution-prior (LeWM
\citep{maes2026leworldmodel}), or action-sensitive (Delta-JEPA \citep{zhang2026deltajepa},
SMWM \citep{ivashkov2026sensorimotor}). And with its standardized simulated environments and planners, stable-worldmodel (SWM) makes world-model results comparable. However, almost all of the existing works see the world through a camera.

Images are a rational default: cheap sensors, dense semantics, and transferable foundation
models \citep{kirillov2023segment,oquab2023dinov2,radford2021learning,schwingshackl2025fewshot}.
But most robotic manipulation is inherently geometric, and pose is its operative quantity
\citep{thalhammer2024challenges,newbury2023grasp}: pick-and-place is specified by a target
 pose of the object, not by an image of the object once placed
\citep{zeng2020transporter,sundermeyer2021contactgraspnet}, and success is
a matter of where things are, not what they look like. An appearance-driven planner, by contrast, needs a goal image to encode, whether rendered, demonstrated, or generated \citep{nair2018visual,du2023unipi,black2024susie}. Latent planners relax this by comparing latents rather than pixels, lowering the bar on goal-image quality, but the comparison is still appearance-based.

Moreover, images are only one of many sensing modalities in robotics. Robots also perceive the world geometrically, and detection on raw range scans has matured accordingly, progressing from 3D bounding boxes \citep{misra2021endtoend} to 6-DoF and parametric object poses \citep{oberweger2026pi3detr, schwingshackl2026piratr}.

\begin{figure}[t]
  \centering
  \includegraphics[width=\linewidth]{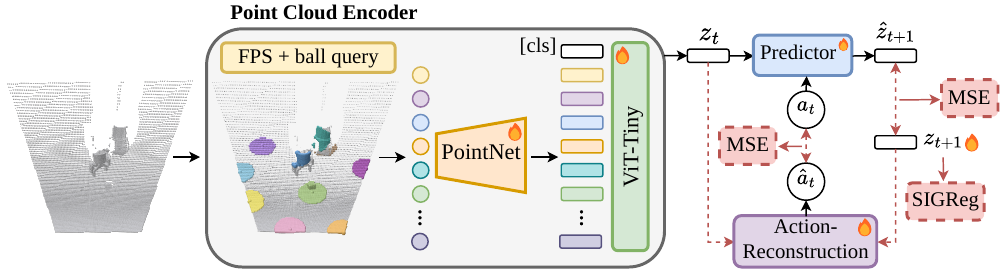}
  \caption{\textbf{Encoder of \plewm{}} and \textbf{\pdjepa{}}: Furthest point sampling (FPS) + radius query (only a few queries are shown), PointNet \citep{qi2017pointnetcvpr} tokenizer, and ViT-Tiny whose class token (CLS) yields $\vz_t$. The next latent $\vz_{t+1}$, from the same encoder, is matched to $\hat{\vz}_{t+1}$ via MSE. SIGReg (\plewm{}) and the Action-Reconstruction-Loss (\pdjepa{}) prevent collapse.}
  \label{fig:arch_pointvit}
\end{figure}

\begin{figure}[t]
  \centering
  \includegraphics[width=\linewidth]{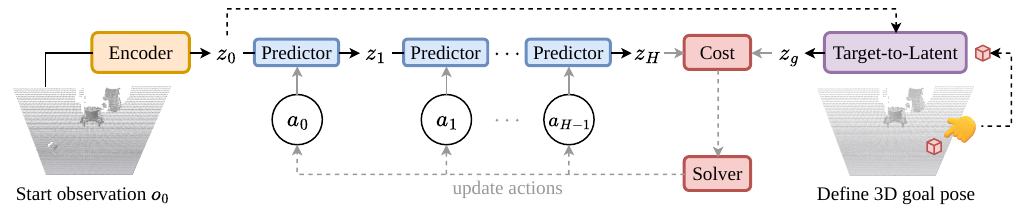}
  \caption{\textbf{Planning toward 3D goal poses.} The target-to-latent module enables specifying a goal directly as a 3D pose. This pose is encoded into a goal latent, which the planner then optimises against, removing the need for a goal observation.}
  \label{fig:arch_target2latent}
\end{figure}

%We therefore want world models that operate directly on geometry,
%so that a planner reasons about the consequences of actions in terms of geometry rather %than
%appearance.
In an effort to make planning with latent action world models more accessible to roboticists, we propose shifting from an appearance-based to a geometry-based paradigm. This is accomplished by (a) using LiDAR-style raycasted point clouds as input for the adapted JEPA architecture (see Figure \ref{fig:arch_pointvit}), and (b) using goal latents derived from 3D target poses as explicit goal specification (Figure~\ref{fig:arch_target2latent}). But this opens up the driving research question behind this work: Does latent planning survive this shift from appearance to point clouds?

Existing 3D world models do not answer it: driving models generate future scans or
occupancy judged by geometric fidelity \citep{zhang2023copilot4d,zheng2024occworld,%
yang2024visual,liu2025listar}, manipulation models decode explicit future geometry 
\citep{huang2026pointworld,peri20263d}, and AD-L-JEPA \citep{zhu2026adljepa} predicts
latents but is action-free. To our knowledge, no prior work learns an
action-conditioned latent-prediction world model on point clouds and plans with it.

We build that missing piece. First, we create the necessary data source by extending the established SWM evaluation platform \citep{maes2026stableworldmodel} with simulated LiDAR-style raycast point clouds as input modality (Section \ref{sec:pc-replication}).
Then, we build action-conditioned world models for point clouds for three JEPA designs: frozen-encoder (DINO-WM-like), distribution-prior (LeWorldModel-like), and action-sensitive (Delta-JEPA-like) (Section \ref{sec:model_architecture}). We add a novel goal-encoding mechanism avoiding goal images or goal point clouds (Section \ref{sec:goal}) and plan with the SWM-MPC.
We isolate each of their respective anti-collapse mechanisms and evaluate pointcloud-based vs. image-based approaches (Section~\ref{sec:experiments}). In Section~\ref{sec:conclusion}, we conclude that comparable results can be achieved with point clouds and therefore that the modality shift from appearance to geometry can be achieved. Our contributions are:

%We investigate another way: retain an explicit
%goal specification, represent it as the 3D pose of a task-relevant object, and
%learn to construct the goal latent the planners above lack without a goal
%observation.

\begin{itemize}

\item \textbf{LiDAR for SWM}: Simulated LiDAR-like point clouds as a new sensor modality for the SWM platform and benchmark.
\item \textbf{\plewm{} and \pdjepa{}}: Point-cloud adaptations of LeWM and Delta-JEPA, which use the anti-collapse approaches distribution-prior and action-sensitive, respectively.
%\item \textbf{\pdjepa{}}: An action-sensitive point-cloud adaptation of Delta-JEPA.
%\item \textbf{\utoniawm{}}: A novel point-cloud JEPA world model built on a frozen Utonia encoder~\citep{zhang2026utonia}.
\item \textbf{\utoniawm{}}: A novel point-cloud JEPA world model built on a frozen Utonia encoder~\citep{zhang2026utonia}, and a matched voxel-feature baseline, \voxwm{}, that isolates the effect of the pretrained features.
\item \textbf{Planning toward 3D targets}: Encoding the goal latent from the current latent and a target pose, replacing goal images.
\item \textbf{Point-cloud vs.\ image-based world models}: A comparative evaluation of all anti-collapse approaches on LiDAR-enabled SWM, showing that point-cloud models can match their image-based counterparts.

  %variant over the , distribution-prior
  %(\plewm{}), and action-sensitive (\pdjepa{}), plus \voxwm{}, a parameter-free
  %geometric ablation that \utoniawm{} outperforms. On the sparse, near-static scans
  %none collapses: positions stay linearly readable from the latents, attention settles
  %on the moving entities, and planning largely withstands sensor degradation never seen
  %in training.
  %\item \textbf{A controlled modality comparison}: a re-sensed
  %\texttt{stable-worldmodel} benchmark where only the observation differs from the
  %image datasets, verified by replay. On it, \plewm{} is statistically
  %equivalent to the image baseline on every benchmark, and \pdjepa{} exceeds
  %it where the most geometry moves (above even its image counterpart's reported,
  %uncontrolled numbers).
\end{itemize}
By providing all models as open weight and source (training and inference) under \url{double-blind.com} we hope to make the classically very image driven world model capabilities accessible to the robotics community interested in exploiting them for LiDAR and pose-directed tasks.
% for point delta jepa we get the highest sucess rate for ogb cube which is the most general 3d dataset, for two room we are comparable

\section{Related Work}
\label{sec:related-work}

\paragraph{JEPA world models.}
Where classical world models learn latent dynamics through reconstruction
\citep{ha2018world,hafner2019learning} or reward prediction \citep{hansen2023tdmpc2},
JEPA-style world models \citep{lecun2022path} learn them by predicting future representations directly, without a decoder or reward.
Trained end-to-end, latent prediction alone collapses, and
the reward-free designs differ in the anti-collapse route they take. DINO-WM
\citep{zhou2025dinowm} predicts the tokens of a frozen DINOv2 \citep{oquab2023dinov2} encoder. PLDM \citep{sobal2025learning} and LeWorldModel
\citep{maes2026leworldmodel} regularize the latent distribution, the latter with a
single term (SIGReg \citep{balestriero2025lejepa}). SMWM
\citep{ivashkov2026sensorimotor} and Delta-JEPA \citep{zhang2026deltajepa}
add an inverse-dynamics loss on consecutive latents or their displacement, so
latents must discriminate between actions. The
SWM platform \citep{maes2026stableworldmodel} consolidates
the first three and their planners into the interface we re-sense. Our \utoniawm{}, \plewm{} and \pdjepa{} carry the three anti-collapse routes
to point clouds. \utoniawm{} takes the frozen-encoder route of DINO-WM with Utonia \citep{zhang2026utonia}, a reusable backbone pretrained on diverse point-cloud data. We distinguish world models from world action models (WAMs)~\citep{ye2026world,li2026causal}, which typically build on video generators and use a learned policy to infer actions directly. Our models instead predict how a given action changes the environment and leave the choice of actions to a planner.

\paragraph{World models on point clouds.}
To the best of our knowledge, none of the existing work plans in latent space. \emph{Manipulation} models regress future geometry: PointWorld \citep{huang2026pointworld} predicts
per-point displacements and drives MPPI on a real arm, 3D Point World Models
\citep{peri20263d} place point completion before the dynamics, and both plan over
explicit predicted point clouds. \emph{Driving} models are generative,
synthesizing future scans or occupancy
\citep{zhang2023copilot4d,zheng2024occworld,yang2024visual,liu2025listar} without
searching over actions. Closest to us, AD-L-JEPA \citep{zhu2026adljepa} brings
JEPA to automotive LiDAR, but as a representation learner. It predicts masked bird's-eye-view features within one scan, so neither time nor actions enter, and its encoder is fine-tuned for object
detection instead of being queried by a planner. Its temporal extension
\citep{zhu2026selfsupervised} forecasts occupancy latents, with SIGReg, but is also
action-free. Outside driving, JEPA has been applied to point clouds without time or actions. Point-JEPA~\citep{saito2025point} learns representations of single objects from coordinates alone, and Locate~3D~\citep{arnaud2025locate} pretrains a 3D-JEPA on scene point clouds enriched with 2D image features.

Beyond this gap, point clouds are a difficult domain for latent prediction, which admits a trivial optimum at temporally constant features \citep{sobal2022joint}, and point clouds are full of such features (further referred to as \emph{slow-feature shortcut}). Height and surface normals follow directly from the coordinates, and \citet{wu2025sonata} show that 3D self-supervised encoders rely on these cues instead of learning semantics, which they call the \emph{geometric shortcut}. In our scans 80--98\% of returns lie on a static ground plane and only 0.3--15\% on anything that moves (Table~\ref{tab:pc-datasets}). The most prominent cues are thus also the most constant ones, promoting the slow-feature shortcut. Whether the mechanisms we study survive this compounding is the empirical question of Section~\ref{sec:experiments}.

\paragraph{Goal specification and planning.}
Sampling-based model predictive control (MPC) is the common planning strategy of latent world models
\citep{hafner2019learning,hansen2023tdmpc2,zhou2025dinowm}, and in the
reward-free setting the goal enters as an \emph{encoded goal observation}
\citep{maes2026stableworldmodel}. Work that improves the search itself \citep{wang2026prism,cheng2026sage,nguyen2026latent} keeps the goal observation, presuming that such an observation exists. FF-JEPA \citep{masip2026ffjepa} removes
goal images by forecasting latent subgoals, but its objective is implicit in the
demonstrations, so the planner cannot be commanded to an arbitrary target. In 3D,
PointWorld \citep{huang2026pointworld} plans toward target positions for a
selected subset of points, a sparse goal cloud with known correspondences that
only a model predicting explicit geometry can evaluate. Our target-to-latent
module keeps an explicit, controllable goal. It is specified by the 3D pose of the task-relevant
object, and constructs the goal latent from that pose and the current latent, giving
latent planners the coordinate interface of geometry-predicting models without
decoding geometry.

\section{Methods}
Section~\ref{sec:pc-replication} discusses how simulated LiDAR is added to the SWM environments. 
Section~\ref{sec:models} builds the three JEPA world-model designs on the resulting point clouds and introduces a complementary module that removes the need for a global goal observation at inference time.

\subsection{Point-cloud sensing and dataset replication}
\label{sec:pc-replication}
Since our work is a direct successor to LeWM, we consider the same environments: OGB-Cube~\citep{park2025ogbench}, Two-Room~\citep{sobal2025learning}, Push-T~\citep{chi2023diffusionpolicy}, and Reacher~\citep{todorov2012mujoco}. To compare observation modalities rather than different datasets, we \emph{re-sense} the released image benchmarks. Specifically, we attach a raycast range sensor to the corresponding simulator, replay every recorded state, and store the resulting point cloud alongside the original observation. Thus, episodes, episode lengths, actions, and goals are identical to those of the image datasets (Table~\ref{tab:pc-datasets-full}).

\begin{figure}
    \centering
    \includegraphics[width=1\linewidth]{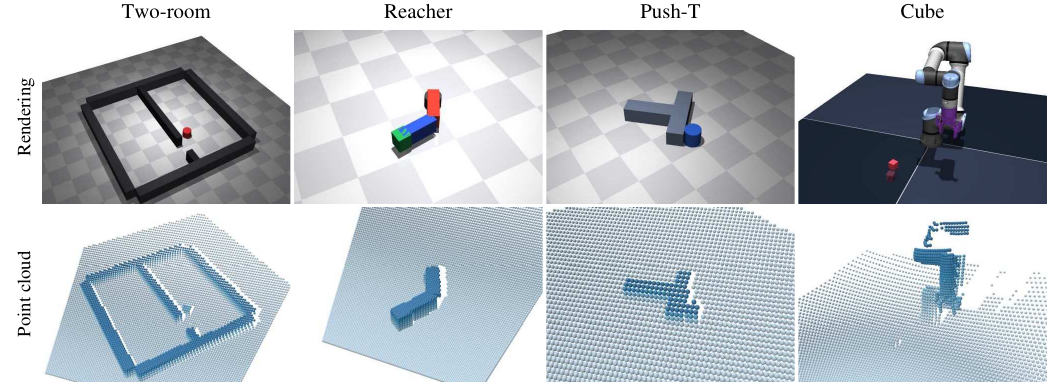}
    \caption{\textbf{From image environments to point cloud environments.} Top: the 3D
    scene. Bottom: the generated point cloud of that same scene, which replaces the image
    as the observation.}
    \label{fig:placeholder}
\end{figure}

\paragraph{Sensor.}
The sensor casts a fixed $100\times100$ grid of rays matched to its mount
camera's field of view, recording each ray's first intersection within a
per-environment maximum range. Returns are expressed in the sensor frame ($x$
forward, $y$ left, $z$ up), and rays intersecting nothing take a sentinel
value, so each frame is a fixed $(10{,}000,3)$ array. The sensor is idealized,
exact intersections, no noise, dropout or sweep distortion, so the comparison
isolates observation geometry. Only the OGBench cube
provides a mountable camera. The others offer a top-down view alone, from
which a scan reduces to a height map. Hence, for these environments, we place the sensor at $45^\circ$
azimuth and elevation, derived from the scene's bounding sphere rather than
tuned, to create more realistic LiDAR scene simulations.

\begin{table}[t]
  \centering
  \caption{\textbf{The four re-sensed datasets.} \emph{Ground share} and \emph{moving
  returns} are the fractions of returns on the ground plane and on non-static
  entities.}
  \label{tab:pc-datasets}
  \footnotesize\setlength{\tabcolsep}{4pt}
  \begin{tabular}{lcccc}
    \toprule
     & OGB-Cube & Two-Room & Push-T & Reacher \\
    \midrule
    Ground share       & $\sim$79\% & $\sim$80\% & $\sim$98\% & $\sim$97\% \\
    Moving returns     & $\sim$15.0\% & $\sim$0.3\% & $\sim$0.5\% & $\sim$2.0\% \\
    \bottomrule
  \end{tabular}
\end{table}

\paragraph{Dataset conversion.}
We set the simulator to each state recorded in the original dataset and cast a LiDAR scan, adding the point cloud alongside the original data. Two-Room and Push-T are not 3D simulators, so we leave each engine unmodified and mirror its state into a lightweight MuJoCo~\citep{todorov2012mujoco} scene that serves only as geometry for the sensor (Appendix~\ref{app:pc-detail}). Dynamics, rewards, and termination are still computed by the original environment. At evaluation time, the same sensor scans each state the planner visits. In the resulting clouds, ground returns dominate and only 0.3--15\% of points fall on moving entities (Table~\ref{tab:pc-datasets}), which poses two distinct challenges for latent prediction (Section~\ref{sec:related-work}).

\subsection{Point cloud world models and planning toward 3D goal poses}
\label{sec:models}

We adopt the offline, reward-free setting of prior JEPA world models
\citep{maes2026leworldmodel}: the training data
is a set of $n$ action-interleaved trajectories
$\gD=\{(o_1,\va_1,\dots,\va_{T_i-1},o_{T_i})\}_{i=1}^{n}$, without rewards or task labels. Each
observation $o_t$ is a point cloud, an unordered point set in $\R^3$ of
frame-varying size, and each action $\va_t$ a continuous control input. At test time
the model must plan actions that reach a specified goal, conventionally given as a
\emph{goal observation} $o_g$ \citep{maes2026stableworldmodel, zhou2025dinowm}. Our target-to-latent module will instead assume training-time annotations $\vc_t$ of low-dimensional 3D targets, the only supervision used in this paper.

\subsubsection{Model architecture}
\label{sec:model_architecture}
All models follow the JEPA template \citep{lecun2022path}: an encoder $f_\theta$ maps each observation to a latent representation, and a predictor $p_\phi$ predicts the next latent from a context of $\ell$ latents and actions,
\begin{equation}
\label{eq:jepa}
\vz_t=f_\theta(o_t)\in\gZ,\qquad
\hat{\vz}_{t+1}=p_\phi(\vz_{t-\ell+1:t},\va_{t-\ell+1:t}).
\end{equation}
Observations are never reconstructed. \plewm{} and \pdjepa{} use the same encoder and predictor architectures and are trained end-to-end, differing only in their objectives. This controlled setup isolates the effect of their respective anti-collapse mechanisms. In contrast, \utoniawm{} does not learn an encoder.

\paragraph{\plewm{} \& \pdjepa{}.} For the end-to-end pair, $f_\theta$ adapts LeWorldModel's pixel encoder to point clouds (Figure~\ref{fig:arch_pointvit}): a Point-BERT-style set tokenizer \citep{yu2022pointbert,zhou2024uni3d} feeds the image model's ViT-Tiny trunk \citep{dosovitskiy2021image}, whose \textsc{CLS} token gives $\gZ=\R^{192}$. The predictor is LeWorldModel's causal action-conditioned transformer \citep{maes2026leworldmodel}. We make three changes to standard recognition practice. Coordinates are normalized using a \emph{fixed} workspace affine rather than per cloud, since per-cloud statistics would remove the absolute object positions that define the task signal. Points are grouped by Euclidean radius \citep{qi2017pointnet} rather than $k$-NN, so each token covers a fixed physical scale. Finally, the encoder uses only LayerNorm, following R3D's fix \citep{hong2026r3d} for the BatchNorm-induced 3D ``scaling paradox'' in control.

\begin{figure}[t]
  \centering
  \includegraphics[width=\linewidth]{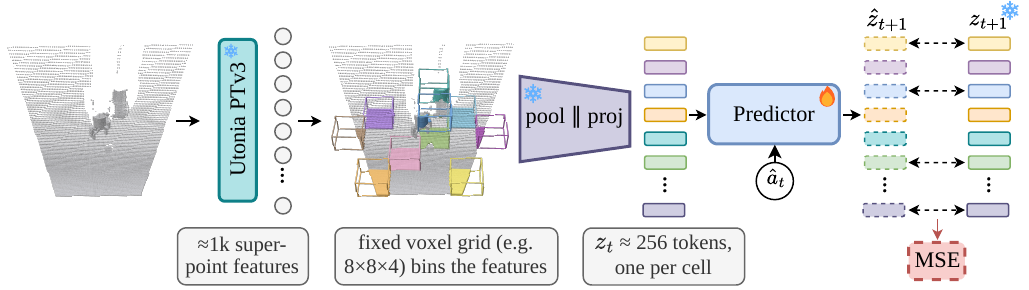}
  \caption{\textbf{\utoniawm{}}: frozen Utonia \citep{zhang2026utonia} encodes the point cloud into super-point features, which a fixed canonical-frame voxel grid bins into ${\approx}256$ tokens (only a few cells shown). Per cell, features are pooled and compressed by a fixed random orthogonal projection to the predictor width. Only the predictor trains, with a token-wise MSE against the frozen targets $\vz_{t+1}$.}
  \label{fig:arch_utoniawm}
\end{figure}

\paragraph{\utoniawm{}.} \utoniawm{} (Figure~\ref{fig:arch_utoniawm}) follows DINO-WM \citep{zhou2025dinowm}: 
a frozen pretrained encoder supplies the state, and only the predictor trains. 
Our frozen encoder is Utonia \citep{zhang2026utonia}, a Point Transformer~V3 \citep{wu2023point} consuming point coordinates alone.
Unlike image patch tokens, Utonia's super-points have no persistent identity across frames, making DINO-WM's token-wise prediction undefined. 
We therefore rotate each point cloud into the gravity-aligned frame expected by Utonia, encode it with the frozen backbone, and pool the resulting features into a fixed voxel grid, yielding ${\approx}256$ tokens.
Each pooled token is then compressed from its 1152 pooled dimensions to the 192-dimensional predictor width by a fixed random orthogonal projection (Appendix~\ref{app:utonia-wm}). The projection is fixed and training-free, while approximately preserving distances by the Johnson--Lindenstrauss lemma \citep{johnson1984extensions}.
These tokens are predicted by DINO-WM's frame-block-causal transformer \citep{zhou2025dinowm}.
Nothing downstream of the backbone is trainable, so the frozen latent space cannot collapse under the prediction loss. 

\paragraph{\voxwm{}.} \voxwm{} replaces the frozen Utonia features with 25 point statistics per voxel (occupancy, count, and per-axis moments; Appendix~\ref{app:vox-wm}), while keeping the same gravity-aligned frame, voxel grid, predictor, and fixed projection to 192 dimensions. Thus, the two models differ only in their voxel features, isolating the contribution of Utonia's learned representation.

\paragraph{Joint training objectives.}
All models minimize the teacher-forced latent prediction loss
\citep{zhou2025dinowm,maes2026leworldmodel}
\begin{equation}
\label{eq:pred}
\Ls_{\mathrm{pred}}(\theta,\phi)=\E_{\gD}\Bigl[\bigl\lVert p_\phi(\vz_{t-\ell+1:t},\va_{t-\ell+1:t})-\vz_{t+1}\bigr\rVert_2^2\Bigr],
\end{equation}
i.e.\ the prediction of \Eqref{eq:jepa} must match the encoded next observation. Trained end-to-end, this loss alone leads to a collapsed solution \citep{sobal2022joint}. 
\plewm{} (P-LeWM) and \pdjepa{} (PD-JEPA) avoid collapse by adding different auxiliary objectives to \Eqref{eq:pred}:
\begin{align}
\Ls_{\text{P-LeWM}}(\theta,\phi)&=\Ls_{\mathrm{pred}}+\lambda\,\Omega_{\mathrm{SIGReg}}(\theta),\label{eq:plewm}\\
\Ls_{\text{PD-JEPA}}(\theta,\phi,\psi)&=\Ls_{\mathrm{pred}}+\beta\,\E_{\gD}\Bigl[\frac{1}{N}\bigl\lVert h_\psi(\vz_{t+N}-\vz_t)-\va_{t:t+N-1}\bigr\rVert_2^2\Bigr].\label{eq:pdjepa}
\end{align}
In \Eqref{eq:plewm}, the SIGReg regularizer
\citep{balestriero2025lejepa,maes2026leworldmodel}, weighted by $\lambda$, drives
the latent distribution toward an isotropic Gaussian, preventing collapse. In \Eqref{eq:pdjepa}, a training-only decoder $h_\psi$, discarded
after training and weighted by $\beta$, must recover the $N{=}5$ executed actions from the change in latent representation, yielding latents that discriminate between actions
\citep{zhang2026deltajepa,ivashkov2026sensorimotor}.
\utoniawm{} needs neither term: only the predictor is trained against the
fixed targets of the frozen encoder, so its representation cannot adapt to the task.

\paragraph{Implementation.}
All models are implemented in PyTorch \citep{paszke2019pytorch} and trained on two A100 GPUs within a 72-hour training budget per run. The shared end-to-end encoder is a ViT-Tiny trunk of ${\approx}5.5$M parameters with a 192-dimensional latent, trained for 10 epochs, whereas \utoniawm{} pairs its frozen ${\approx}137$M-parameter backbone with a predictor trained for 100 epochs on precomputed features. All models step at frameskip 5 and plan from a context of $\ell=3$ past latents. We reimplemented image Delta-JEPA and DINO-WM from their official implementations, adjusting batch size and learning rate to the 72-hour budget. For LeWM, we evaluate the released checkpoint. Under our evaluation protocol all three land at or near their published success rates (Appendix~\ref{app:baseline-fidelity}), with retrained Delta-JEPA matching or exceeding its published numbers on every benchmark. More implementation details are provided in Appendix~\ref{app:impl}.

\subsubsection{Planning toward 3D goal poses}
\label{sec:goal}
As in the pixel models~\citep{maes2026leworldmodel, zhou2025dinowm}, we plan with receding-horizon MPC in latent space, using the cross-entropy method (CEM) to find action sequences of horizon $H$ that minimize $\lVert \hat{\vz}_{t+H} - \vz_g \rVert_2^2$. Here $\hat{\vz}_{t+H}$ is the autoregressive rollout of the predictor in \Eqref{eq:jepa}, and $\vz_g$ is a goal latent, conventionally an encoded goal observation $\vz_g = f_\theta(o_g)$. In robotics, however, goals are usually given as target placements of the robot or objects. To avoid requiring a scan of an unvisited state, we instead specify the goal as a low-dimensional 3D target $\vc_g \in \gC$ containing only task-relevant coordinates: the cube position in OGB-Cube, the agent position in Two-Room, the agent and block positions plus the block's planar angle in Push-T, and the joint positions in Reacher.

With $f_\theta$ frozen, we then train a target-to-latent mapping $m_\xi : \gC \times \gZ \to \gZ$ on pairs of frames from the same trajectory, separated by a random offset $k \le k_{\max}$:
\begin{equation}
\label{eq:goal}
\Ls_{\mathrm{goal}}(\xi) = \E_{(o_t,\, \vc_{t+k},\, o_{t+k}) \sim \gD,\; k \sim \mathcal{U}\{1,\dots,k_{\max}\}} \Bigl[ \bigl\lVert m_\xi(\vc_{t+k}, \vz_t) - \vz_{t+k} \bigr\rVert_2^2 \Bigr],
\end{equation}
where $\vz_t = f_\theta(o_t)$ and $\vz_{t+k} = f_\theta(o_{t+k})$. A target alone cannot determine a latent, because $f_\theta$ encodes the whole scene while $\vc_g$ fixes only a few coordinates. The current latent $\vz_t$ supplies most of the missing scene content. Some ambiguity remains, since parts of the scene can change over $k$ steps, and a regression model then predicts the mean of the plausible latents, which need not match any encoder output (Appendix~\ref{app:goal-ambiguity}). We therefore compare two independent choices in Table~\ref{tab:target2latent}. First, whether $m_\xi$ is conditioned on $\vz_t$. Second, its architecture: a residual MLP over Fourier-featurized target coordinates trained with \Eqref{eq:goal}, or a shortcut model~\citep{frans2025one} that samples a latent in one step and is trained with its own flow-matching objective on the same pairs.

At test time, $\vz_g = m_\xi(\vc_g, \vz_t)$ or $\vz_g = m_\xi(\vc_g)$ replaces the encoded goal observation in the planning objective.
The world model and planner are unchanged. Planning details and hyperparameters are listed in Appendix~\ref{app:planning}, Architectures and training details are in Appendix~\ref{app:target-to-latent}, and Appendix~\ref{app:goal-ambiguity} discusses rendering a synthetic goal scan as an alternative.
\section{Experiments}
\newcommand{\sd}[1]{\ensuremath{_{\pm#1}}}
 \label{sec:experiments}

All models are trained on the datasets of Section~\ref{sec:pc-replication}, where each image episode has a matching point cloud episode, so both modalities share the same underlying scenarios. We evaluate goal-conditioned planning with receding-horizon CEM from the SWM solver layer~\citep{maes2026stableworldmodel}, using 10 seeds of 50 episodes each and the same fixed start--goal pairs for all models.
%
%\subsection{Baselines}
%
\begin{table}[t]
  \centering
  \caption{\textbf{Planning success rate (\%)}. Mean$_{\pm\text{sd}}$ over 10 evaluation seeds of 50 episodes each
  (sd across seeds, here and throughout),
  with CEM behind the SWM solver~\citep{maes2026stableworldmodel} on
  identical fixed start--goal pairs. Bold marks the best model within each family.}
  \label{tab:main}
  \begin{tabular}{@{}lllllll@{}}
    \toprule
    \textbf{Family} & \textbf{Obs.} & \textbf{Method} & \textbf{Two-Room} & \textbf{Reacher} & \textbf{Push-T} & \textbf{OGB-Cube} \\
    \midrule
      &  & Random & 25.2\sd{4.1} & 10.8\sd{4.0} & 2.4\sd{1.3} & 46.0\sd{8.5} \\
    \midrule
    \multirow{2}{*}{SIGReg}
      & Image & LeWM & 85.6\sd{5.7} & \textbf{82.0\sd{5.6}} & \textbf{86.4\sd{5.2}} & \textbf{69.8\sd{9.3}} \\
      & Points & Point-LeWM & \textbf{87.0\sd{4.4}} & 80.4\sd{4.6} & 83.6\sd{3.4} & 66.0\sd{5.9} \\
    \midrule
    \multirow{2}{*}{Act.-reconst.}
      & Image & Delta-JEPA
        & 99.8\sd{0.6} & \textbf{86.0\sd{3.3}} & \textbf{94.2\sd{2.7}} & 80.2\sd{6.0} \\
      & Points & Point-Delta-JEPA & \textbf{100.0\sd{0.0}} & 77.6\sd{4.4} & 70.8\sd{7.7} & \textbf{83.4\sd{3.8}} \\
    \midrule
    \multirow{3}{*}{Frozen enc.}
      & Image & DINO-WM
        & \textbf{99.8\sd{0.6}} & \textbf{77.0\sd{4.8}} & \textbf{76.0\sd{9.5}} & \textbf{78.6\sd{6.7}} \\
      & Points & Utonia-WM & 94.0\sd{1.9} & 71.2\sd{3.7} & 46.6\sd{6.6} & 71.2\sd{7.7} \\
      & Points & Vox-WM & 83.8\sd{5.2} & 71.8\sd{6.7} &  55.8\sd{4.9} & 65.2\sd{9.1} \\
    \bottomrule
  \end{tabular}
\end{table}
Table~\ref{tab:main} shows that every point cloud world model plans far above Random despite moving fractions of only 0.3--15\%, so neither 
the geometric shortcut nor the slow-feature shortcut of Section~\ref{sec:related-work} dominates. Significance tests are paired over seeds (Appendix~\ref{app:stats}). Inference and planning times of the evaluated models are analyzed in Appendix~\ref{app:compute}.

\paragraph{SIGReg \& action reconstruction.} \plewm{} does not differ significantly from LeWM on any benchmark, and the largest deficit is 3.8 points on OGB-Cube.
\pdjepa{} matches image Delta-JEPA, retrained with the same time budget \citep{zhang2026deltajepa},
on Two-Room and on OGB-Cube, where it attains the best result of
Table~\ref{tab:main}. On Reacher and Push-T, it trails significantly, by 8.4
and 23.4 points. The point cloud input alone cannot explain these gaps,
since \plewm{} shows none in the same environments. Compared with \plewm{},
\pdjepa{} gains 13.0 and 17.4 points on Two-Room and OGB-Cube but loses 12.8
on Push-T. Probing suggests that these differences reflect which task-relevant information each objective preserves (Section~\ref{sec:probing}). The results therefore show that the choice of anti-collapse objective depends on the task.

\paragraph{Frozen encoder.} \utoniawm{} transfers the frozen-encoder recipe to
point clouds but trails DINO-WM on every benchmark, most on Push-T
($-29.4$, versus $-5.8$ to $-7.4$ elsewhere). \voxwm{} replaces only its
pretrained features with voxel statistics. Utonia features help on Two-Room
($+10.2$) and OGB-Cube ($+6.0$), make little difference on Reacher ($-0.6$),
and hurt on Push-T ($-9.2$). In both scans, 97--98\% of returns lie on the
ground plane beneath low, thin entities (Table~\ref{tab:pc-datasets}), so
voxel occupancy and moments already capture most of the geometry. Pretrained
features help only where genuine 3D structure matters, such as the walls of
Two-Room and the arm and cube of OGB-Cube. In Appendix \ref{app:utonia-wm}, we provide an ablation of \utoniawm{}'s design choices.

\begin{figure}[!b]
  \centering
  \includegraphics[width=\linewidth]{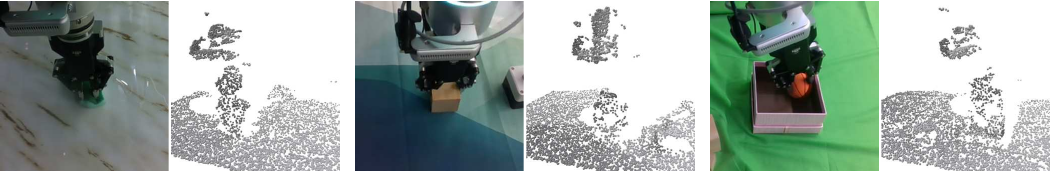}\\[0.4em]
  \includegraphics[width=\linewidth]{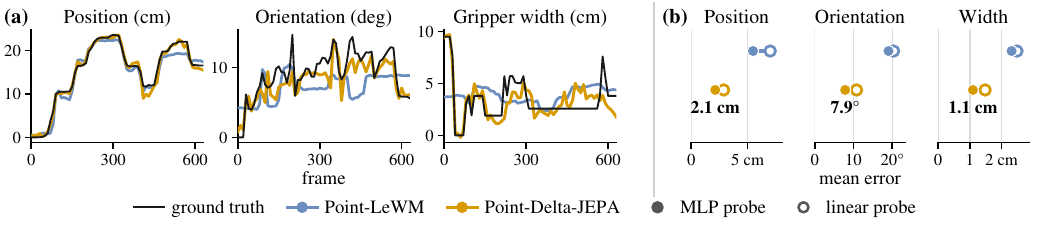}
  \vspace{-1.4em}  % pull the caption towards the figure
  \caption{\textbf{Probing on real-world RH20T data.}
  \emph{Top:} three camera and point cloud samples. The models only see the 3D positions of
  10,000 points per cloud. Images and point colors are for visualization only.
  \emph{Bottom:} what the latents encode about the gripper.
  (a)~For each frame of a held-out grasping episode, the true gripper state is compared with the state an MLP probe decodes from that frame's latent. Additional samples are visible in Figure~\ref{fig:rh20t_eight_samples}.
  (b)~Mean probe error over all 400 held-out episodes (filled: MLP probe, hollow: linear probe).}
  \label{fig:rh20t}
\end{figure}

\subsection{Evaluating Physical World Understanding}
In this subsection, we examine what the latent spaces of \plewm{} and \pdjepa{} encode, how well physical state can be decoded from them, and how robust planning is to degraded input point clouds. Finally, we evaluate these models with physical latent probing on a real-world dataset.

\paragraph{Physical latent probing.}
\label{sec:probing}
We regress the privileged simulator state from the 192-dimensional latent used by the CEM cost, training linear (ridge) and MLP probes on episode-disjoint splits (Appendix~\ref{app:probing}). Both models encode \emph{positions} well in every environment: linear probes reach $R^2 \geq 0.94$ and MLP probes $R^2 \geq 0.98$ (Tables~\ref{tab:probe-pusht}--\ref{tab:probe-cube}). 
However, they differ in how well they encode the Push-T block \emph{angle} (Table~\ref{tab:probe-pusht}), which is linearly decodable from \plewm{} ($R^2 = 0.89$) but much less so from \pdjepa{} ($R^2 = 0.66$). An MLP recovers the angle from both models ($R^2 = 0.99$ and $0.96$), so \pdjepa{} retains most of this information but encodes it non-linearly. 
Since Push-T success requires matching the block's orientation as well as its position, this difference in angle encoding is consistent with the 13-point gap in success rate in favor of \plewm{}.
On OGB-Cube, \pdjepa{} encodes joint \emph{velocity} substantially better than \plewm{} (MLP $R^2 = 0.63$ vs.\ $0.10$; Table~\ref{tab:probe-cube}). This aligns with its action-reconstruction objective and its stronger planning performance on this environment.

\paragraph{Point cloud perturbations.}
We degrade the evaluation cloud with along-ray range noise ($\sigma \in \{0.25, 0.5\}\%$ of the scan's bounding-box diagonal) and random dropout ($25$ or $50\%$ of returns), applied either to the observation alone or to both observation and goal (Table~\ref{tab:ablation-compare}, Appendix~\ref{app:degredation}). Degrading the goal as well lowers success by a further 6\% (relative) on average. We discuss this setting in Section~\ref{sec:ex_t2l} and report clean-goal results here. On the sparse scenes, \plewm{} is more robust. Under $0.25\%$ noise on Two-Room, it retains 80\% of its clean success rate versus 43\% for \pdjepa{}. Under $50\%$ dropout, it retains 99\% on Two-Room and 95\% on Reacher, versus 64\% and 87\% for \pdjepa{}. Attention maps (Appendix~\ref{app:attention}) show that both encoders still attend to moving entities that make up only $0.3\%$ of returns, although \pdjepa{} places more attention on the ground and table. Push-T fails under noise for both models (success below 5\%): with 98\% of returns on the ground plane, the noise overwhelms the thin pusher and block. OGB-Cube is barely affected (98--103\% of clean success under noise, 99--102\% under dropout), possibly because around $15\%$ of its returns come from moving objects, making them easier to separate from the table.

\begin{figure}[t]
    \centering
    \includegraphics[width=1\linewidth]{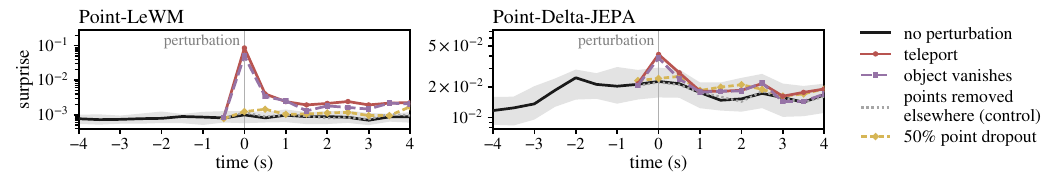}
    \caption{\textbf{Latent surprise on RH20T point clouds.} Median 1-step prediction error
over 32 held-out episodes. From $t=0$ on, an object on the table is moved (teleport) or removed (vanishes), the same number of points is removed elsewhere (control), or half of all points are dropped. Both models react to the physical changes. An example is shown in Figure~\ref{fig:surprise_frames_rh20t}.}
    \label{fig:surprise_rh20t}
\end{figure}
\begin{figure}[!b]
  \centering
  \includegraphics[width=\linewidth]{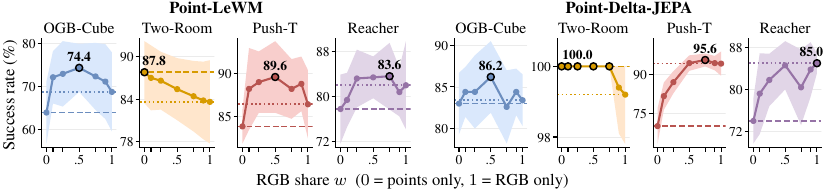}
  \vspace{-1.4em}  % pull the caption towards the figure
  \caption{Cost-level fusion of the point-cloud and RGB world models at planning time. Success rate (mean $\pm$ one standard deviation) over the RGB share $w$, with dashed and dotted lines marking the point-cloud model alone ($w=0$) and the RGB model alone ($w=1$). Outlined markers give the best $w$. Y-axes differ per panel.}
  \label{fig:fusion}
\end{figure}

\paragraph{Real-World Evaluation}
All experiments so far use simulated point clouds, which are complete, noise-free, and come with exact states. Real depth sensors, however, produce clouds that are sparse, noisy, and partly occluded, so it is not clear whether the latents still capture the robot's state. To test this, we train Point-LeWM and Point-Delta-JEPA exclusively on raw point clouds from 1,999 real teleoperated episodes of the RH20T dataset \citep{fang2024rh20t}, recorded with a single RGB-D camera (Figure~\ref{fig:rh20t}, top). Both models keep the same architectures and hyperparameters as in simulation, so nothing is tuned for the real data. We then freeze the encoders and probe their latents for the gripper state on 400 held-out episodes. Point-Delta-JEPA's latents encode the full gripper state: from a single frame, an MLP probe recovers the gripper position to within 2.1 cm, its orientation to within 7.9°, and its opening width to within 1.1 cm, and the decoded state follows the true motion over an episode (Figure~\ref{fig:rh20t}, bottom). Point-LeWM encodes the position (5.5 cm) but hardly the orientation or width. 

Beyond the encoder, the predictor also captures how the scene evolves. When an object on the table is teleported or removed, the latent prediction error rises sharply, while dropping half of all points or removing random patches barely changes it (Figure~\ref{fig:surprise_rh20t}). Although probing and surprise scores do not guarantee planning success, this shows that our JEPA pipeline transfers to real point clouds without model changes, opening the way to planning on real robots. 
More details about the data pipeline, full results, and the experiments can be found in Appendix~\ref{app:rh20t}.

\subsection{Multi-Modal Planning}
\label{sec:multi-modal-planning}
We test whether planning with two world models of different modalities improves success over a single model, both to show the potential of combining world models and to provide a simple baseline for future research. Each model is trained on its own modality only and scores the candidate action sequences with its own latent rollouts, giving the costs $c_{\text{points}}$ and $c_{\text{RGB}}$. Planning then proceeds as usual on the shared cost $c = (1-w)\,\tilde c_{\text{points}} + w\,\tilde c_{\text{RGB}}$, where $\tilde c$ is the per-iteration z-scored cost, $w=0$ is the point-cloud model alone, and $w=1$ is the RGB model alone. Figure~\ref{fig:fusion} (and Table~\ref{tab:fusion} in the Appendix) show the results for \plewm{} and \pdjepa{}, each paired with its RGB counterpart. In five of eight configurations, a combination reaches the highest score, and in the other three, it stays within one standard deviation of the best single model. Most notably, the combination significantly beats both single models on OGB-Cube for both pairs and on Push-T for \plewm{}.

\begin{table}[t]
\centering
\caption{\textbf{Planning success rate (\%) toward 3D targets.} Mean$_{\pm\text{sd}}$ over 10 seeds (50 episodes
each). Target encoders (P-LeWM/PD-JEPA abbreviate \plewm{}/\pdjepa{}) map the 3D target to a
goal latent. $\vz_t$ marks whether the current latent is an additional input. \emph{Goal cloud}
encodes the stored goal cloud (reference). Best per column in bold, separately with and without $\vz_t$.}
\label{tab:target2latent}
\small
\resizebox{\textwidth}{!}{%
\begin{tabular}{llcccccccc}
\toprule
& & \multicolumn{2}{c}{\textbf{Two-Room}} & \multicolumn{2}{c}{\textbf{Reacher}} & \multicolumn{2}{c}{\textbf{Push-T}} & \multicolumn{2}{c}{\textbf{OGB-Cube}} \\
\cmidrule(lr){3-4} \cmidrule(lr){5-6} \cmidrule(lr){7-8} \cmidrule(lr){9-10}
\textbf{Encoder} & $\vz_t$ & P-LeWM & PD-JEPA & P-LeWM & PD-JEPA & P-LeWM & PD-JEPA & P-LeWM & PD-JEPA \\
\midrule
MLP      & \ding{55}  & 87.4 \tiny$\pm$ 4.8 & 100.0 \tiny$\pm$ 0.0 & \textbf{77.4 \tiny$\pm$ 5.3} & \textbf{80.0 \tiny$\pm$ 5.5} & \textbf{85.0 \tiny$\pm$ 3.6} & \textbf{71.8 \tiny$\pm$ 7.1} & \textbf{65.2 \tiny$\pm$ 6.5} & 72.0 \tiny$\pm$ 5.2 \\
Shortcut & \ding{55}  & \textbf{87.6 \tiny$\pm$ 4.8} & 100.0 \tiny$\pm$ 0.0 & 73.6 \tiny$\pm$ 6.7 & 72.8 \tiny$\pm$ 4.5 & 83.2 \tiny$\pm$ 4.4 & 71.4 \tiny$\pm$ 6.8 & 61.6 \tiny$\pm$ 5.7 & \textbf{78.8 \tiny$\pm$ 5.3} \\
\cmidrule(l{0.6em}r{0.6em}){1-10}
MLP      & \checkmark & 88.4 \tiny$\pm$ 4.3 & 100.0 \tiny$\pm$ 0.0 & \textbf{77.0 \tiny$\pm$ 4.2} & \textbf{77.8 \tiny$\pm$ 4.9} & 83.8 \tiny$\pm$ 5.0 & 70.8 \tiny$\pm$ 7.4 & 66.0 \tiny$\pm$ 7.8 & 82.4 \tiny$\pm$ 5.9 \\
Shortcut & \checkmark & \textbf{88.6 \tiny$\pm$ 3.9} & 100.0 \tiny$\pm$ 0.0 & 73.6 \tiny$\pm$ 5.2 & 71.2 \tiny$\pm$ 4.9 & \textbf{84.8 \tiny$\pm$ 2.3} & \textbf{73.6 \tiny$\pm$ 5.8} & \textbf{68.8 \tiny$\pm$ 8.3} & \textbf{83.2 \tiny$\pm$ 5.0} \\
\midrule
\multicolumn{2}{l}{\emph{Goal cloud}} & 87.2 \tiny$\pm$ 4.1 & 100.0 \tiny$\pm$ 0.0 & 78.8 \tiny$\pm$ 4.6 & 76.8 \tiny$\pm$ 2.3 & 84.8 \tiny$\pm$ 4.1 & 71.2 \tiny$\pm$ 5.0 & 65.4 \tiny$\pm$ 7.5 & 84.2 \tiny$\pm$ 4.2 \\
\bottomrule
\end{tabular}%
}
\end{table}

\subsection{Target-to-latent module results}
\label{sec:ex_t2l}
Table~\ref{tab:target2latent} compares planning toward a goal latent predicted by the target-to-latent module (Section~\ref{sec:models}) with planning toward the encoded goal cloud. In every column, at least one encoder conditioned on the current latent $\vz_t$ matches the goal-cloud reference within one standard deviation, showing that a low-dimensional target pose can replace the goal observation without reducing success. As predicted in Section~\ref{sec:goal}, $\vz_t$ matters only when the target leaves some moving entities unconstrained. On Two-Room, Reacher, and Push-T, the target fixes all moving entities, so the unconditioned MLP already matches the reference. On OGB-Cube, where the arm configuration remains free, dropping $\vz_t$ costs up to 10 points (\pdjepa{} with the MLP encoder). Neither encoder is consistently better. The shortcut is best or tied in six of the eight conditioned columns, with a maximum difference of 2.8 points, while the MLP is clearly better for \pdjepa{} on Reacher (77.8 vs.\ 71.2). 

A target pose also has two advantages over a goal cloud. It is unaffected by sensor noise, so when the deployed sensor is noisier than during training, only the observation degrades, and planning stays more robust (Table~\ref{tab:ablation-mlp}). It also generalizes beyond the trained goal distribution, matching planning toward a cloud rendered at the specified state (Appendix~\ref{sec:cube_grid}).

\section{Conclusion}
\label{sec:conclusion}

\paragraph{Limitations.} \utoniawm{} plans as slow as DINO-WM. Its fixed grid pooling and random projection force a token structure onto the model. The target-to-latent module needs training-time target annotations. Planning evaluation is confined to four simulated scenes with an idealized sensor.

\paragraph{Conclusion.} 
Our evaluations show that world models are not limited to image observations and that they can also operate effectively on LiDAR-style point clouds. This holds despite two known trivial shortcuts that can cause point-cloud representations to degenerate: the geometric shortcut and the slow-feature shortcut. 
On OGB-Cube, the most representative of the evaluated 3D environments, action reconstruction (Point-Delta-JEPA) proves to be the most effective anti-collapse approach. Our analysis explains why. Object positions are almost perfectly linearly decodable from the learned representations, attention concentrates on the task-relevant entities, and planning remains robust under heavy dropout not seen during training. The main remaining failure mode is range noise on thin structures. 
Additionally, for planning, we provide a goal-pose encoder that offers a more natural interface for robotic manipulation than encoding an image of the desired end state.
Taken together, these results establish the first geometry-driven, rather than image-based, approach to latent world-model-based planning.
\newpage

\subsection*{Acknowledgments}
We are grateful to Robin Schiewer for reviewing an earlier draft of this work and for his valuable suggestions. The computational results presented have been achieved using the Vienna Scientific Cluster (VSC).

\bibliography{iclr2027_conference}

@misc{balestriero2025lejepa,
  title = {{LeJEPA}: Provable and Scalable Self-Supervised Learning Without the Heuristics},
  author = {Balestriero, Randall and LeCun, Yann},
  year = {2025},
  eprint = {2511.08544},
  archivePrefix = {arXiv}
}

@misc{cheng2026sage,
  title = {{SAGE}: Subgoal-Conditioned Action Generation for Latent World Model Planning},
  author = {Cheng, Letian and Zhang, Qi and Wang, Yisen},
  year = {2026},
  eprint = {2607.17973},
  archivePrefix = {arXiv}
}

@misc{ha2018world,
  title = {World Models},
  author = {Ha, David and Schmidhuber, Jürgen},
  year = {2018},
  eprint = {1803.10122},
  archivePrefix = {arXiv},
  doi = {10.5281/zenodo.1207631}
}

@inproceedings{hafner2019learning,
  title = {Learning Latent Dynamics for Planning from Pixels},
  author = {Hafner, Danijar and Lillicrap, Timothy and Fischer, Ian and Villegas, Ruben and Ha, David and Lee, Honglak and Davidson, James},
  year = {2019},
  booktitle = {ICML}
}

@inproceedings{hansen2023tdmpc2,
  title = {{TD-MPC2}: Scalable, Robust World Models for Continuous Control},
  author = {Hansen, Nicklas and Su, Hao and Wang, Xiaolong},
  year = {2024},
  booktitle = {ICLR}
}

@misc{huang2026pointworld,
  title = {{PointWorld}: Scaling {3D} World Models for In-The-Wild Robotic Manipulation},
  author = {Huang, Wenlong and Chao, Yu-Wei and Mousavian, Arsalan and Liu, Ming-Yu and Fox, Dieter and Mo, Kaichun and Fei-Fei, Li},
  year = {2026},
  eprint = {2601.03782},
  archivePrefix = {arXiv}
}

@misc{ivashkov2026sensorimotor,
  title = {Sensorimotor World Models: Perception for Action via Inverse Dynamics},
  author = {Ivashkov, Petr and Balestriero, Randall and Schölkopf, Bernhard},
  year = {2026},
  eprint = {2606.20104},
  archivePrefix = {arXiv}
}

@misc{lecun2022path,
  title = {A Path Towards Autonomous Machine Intelligence},
  author = {LeCun, Yann},
  year = {2022},
  url = {https://openreview.net/forum?id=BZ5a1r-kVsf}
}

@misc{liu2025listar,
  title = {{LiSTAR}: Ray-Centric World Models for {4D} {LiDAR} Sequences in Autonomous Driving},
  author = {Liu, Pei and Wang, Songtao and Zhang, Lang and Peng, Xingyue and Lyu, Yuandong and Deng, Jiaxin and Lu, Songxin and Ma, Weiliang and Zhang, Xueyang and Zhan, Yifei and Lang, XianPeng and Ma, Jun},
  year = {2025},
  eprint = {2511.16049},
  archivePrefix = {arXiv}
}

@misc{maes2026leworldmodel,
  title = {{LeWorldModel}: Stable End-to-End Joint-Embedding Predictive Architecture from Pixels},
  author = {Maes, Lucas and Le Lidec, Quentin and Scieur, Damien and LeCun, Yann and Balestriero, Randall},
  year = {2026},
  eprint = {2603.19312},
  archivePrefix = {arXiv}
}

@misc{maes2026stableworldmodel,
  title = {stable-worldmodel: A Platform for Reproducible World Modeling Research and Evaluation},
  author = {Maes, Lucas and Le Lidec, Quentin and Facury, Luiz and Massaudi, Nassim and Chaurasia, Ayush and Capuano, Francesco and Gao, Richard and Gillin, Taj and Haramati, Dan and Scieur, Damien and LeCun, Yann and Balestriero, Randall},
  year = {2026},
  eprint = {2605.21800},
  archivePrefix = {arXiv}
}

@misc{masip2026ffjepa,
  title = {{FF-JEPA}: Long-Horizon Planning in World Models with Latent Planners},
  author = {Masip, Sergi and Swinnen, Jonathan and Hu, Yutong and Detry, Renaud and Tuytelaars, Tinne},
  year = {2026},
  eprint = {2606.09311},
  archivePrefix = {arXiv}
}

@misc{nguyen2026latent,
  title = {Latent Geometry Beyond Search: Amortizing Planning in World Models},
  author = {Nguyen, Hoang and Xu, Xiaohao and Huang, Xiaonan},
  year = {2026},
  eprint = {2605.08732},
  archivePrefix = {arXiv}
}

@inproceedings{oberweger2026pi3detr,
  title = {{PI3DETR}: Parametric Instance Detection of {3D} Point Cloud Edges With a Geometry-Aware {3DETR}},
  author = {Oberweger, Fabio F. and Schwingshackl, Michael and Staderini, Vanessa},
  year = {2026},
  booktitle = {3DV},
  doi = {10.1109/3DV69130.2026.00052}
}

@misc{peri20263d,
  title = {{3D} Point World Models: Point Completion Enables More Accurate Dynamics Learning},
  author = {Peri, Skand and Nguyen, Hung and Kim, Chanho and Li, Fuxin and Lee, Stefan},
  year = {2026},
  eprint = {2607.00148},
  archivePrefix = {arXiv}
}

@inproceedings{schwingshackl2025fewshot,
  title = {Few-shot Structure-Informed Machinery Part Segmentation with Foundation Models and Graph Neural Networks},
  author = {Schwingshackl, Michael and Oberweger, Fabio Francisco and Murschitz, Markus},
  year = {2025},
  booktitle = {WACV},
  doi = {10.1109/WACV61041.2025.00200}
}

@inproceedings{schwingshackl2026piratr,
  title = {{PIRATR}: Parametric Object Inference for Robotic Applications with Transformers in {3D} Point Clouds},
  author = {Schwingshackl, Michael and Oberweger, Fabio F. and Niedermeyer, Mario and Huemer, Johannes and Murschitz, Markus},
  year = {2026},
  booktitle = {ICRA}
}

@inproceedings{sobal2022joint,
  title = {Joint Embedding Predictive Architectures Focus on Slow Features},
  author = {Sobal, Vlad and {S V}, Jyothir and Jalagam, Siddhartha and Carion, Nicolas and Cho, Kyunghyun and LeCun, Yann},
  year = {2022},
  booktitle = {NeurIPS Workshop on Self-Supervised Learning: Theory and Practice}
}

@inproceedings{sobal2025learning,
  title = {Learning from Reward-Free Offline Data: A Case for Planning with Latent Dynamics Models},
  author = {Sobal, Vlad and Zhang, Wancong and Cho, Kyunghyun and Balestriero, Randall and Rudner, Tim G. J. and LeCun, Yann},
  year = {2025},
  booktitle = {NeurIPS}
}

@misc{wang2026prism,
  title = {{PRISM}: {PRior-guided} Imagination Sampling in world Models},
  author = {Wang, Yuhai and Xia, Jiawei and Zhou, Rongxuan and Hu, Xiao and Shi, Yongliang and Du, Jing and Ye, Yang},
  year = {2026},
  eprint = {2606.07974},
  archivePrefix = {arXiv}
}

@inproceedings{wu2023point,
  title = {Point Transformer {V3}: Simpler, Faster, Stronger},
  author = {Wu, Xiaoyang and Jiang, Li and Wang, Peng-Shuai and Liu, Zhijian and Liu, Xihui and Qiao, Yu and Ouyang, Wanli and He, Tong and Zhao, Hengshuang},
  year = {2023},
  booktitle = {CVPR}
}

@inproceedings{wu2025sonata,
  title = {Sonata: Self-Supervised Learning of Reliable Point Representations},
  author = {Wu, Xiaoyang and DeTone, Daniel and Frost, Duncan and Shen, Tianwei and Xie, Chris and Yang, Nan and Engel, Jakob and Newcombe, Richard and Zhao, Hengshuang and Straub, Julian},
  year = {2025},
  booktitle = {CVPR}
}

@inproceedings{yang2024visual,
  title = {Visual Point Cloud Forecasting enables Scalable Autonomous Driving},
  author = {Yang, Zetong and Chen, Li and Sun, Yanan and Li, Hongyang},
  year = {2024},
  booktitle = {CVPR},
  doi = {10.1109/CVPR52733.2024.01390}
}

@inproceedings{zhang2023copilot4d,
  title = {{Copilot4D}: Learning Unsupervised World Models for Autonomous Driving via Discrete Diffusion},
  author = {Zhang, Lunjun and Xiong, Yuwen and Yang, Ze and Casas, Sergio and Hu, Rui and Urtasun, Raquel},
  year = {2024},
  booktitle = {ICLR}
}

@misc{zhang2026deltajepa,
  title = {{Delta-JEPA}: Learning Action-Sensitive World Models via Latent Difference Decoding},
  author = {Zhang, Zhenghao and Wang, Yuanxiang and Guan, Zhenyu and Yang, Yujia and Shi, Bingkang and Zong, Tianyu and Yi, Hongzhu and Chao, Guoqing and Chen, Xingchen and Yang, Tiankun and Bao, Chenxi and Yu, Tao and Zhou, Jingjing and Xu, Jungang},
  year = {2026},
  eprint = {2606.31232},
  archivePrefix = {arXiv}
}

@inproceedings{zhang2026utonia,
  title = {Utonia: Toward One Encoder for All Point Clouds},
  author = {Zhang, Yujia and Wu, Xiaoyang and Yang, Yunhan and Fan, Xianzhe and Li, Han and Zhang, Yuechen and Huang, Zehao and Wang, Naiyan and Zhao, Hengshuang},
  year = {2026},
  booktitle = {ICML}
}

@inproceedings{zheng2024occworld,
  title = {{OccWorld}: Learning a {3D} Occupancy World Model for Autonomous Driving},
  author = {Zheng, Wenzhao and Chen, Weiliang and Huang, Yuanhui and Zhang, Borui and Duan, Yueqi and Lu, Jiwen},
  year = {2024},
  booktitle = {ECCV},
  doi = {10.1007/978-3-031-72624-8\_4}
}

@inproceedings{zhou2025dinowm,
  title = {{DINO-WM}: World Models on Pre-trained Visual Features enable Zero-shot Planning},
  author = {Zhou, Gaoyue and Pan, Hengkai and LeCun, Yann and Pinto, Lerrel},
  year = {2025},
  booktitle = {ICML}
}

@inproceedings{zhu2026adljepa,
  title = {Self-Supervised Representation Learning with Joint Embedding Predictive Architecture for Automotive {LiDAR} Object Detection},
  author = {Zhu, Haoran and Dong, Zhenyuan and Topollai, Kristi and Sha, Beiyao and Choromanska, Anna},
  year = {2026},
  booktitle = {AAAI},
  doi = {10.1609/aaai.v40i16.38402}
}

@misc{zhu2026selfsupervised,
  title = {Self-Supervised {JEPA-based} World Models for {LiDAR} Occupancy Completion and Forecasting},
  author = {Zhu, Haoran and Choromanska, Anna},
  year = {2026},
  eprint = {2602.12540},
  archivePrefix = {arXiv}
}

@inproceedings{dosovitskiy2021image,
  title = {An Image Is Worth 16x16 Words: Transformers for Image Recognition at Scale},
  author = {Dosovitskiy, Alexey and Beyer, Lucas and Kolesnikov, Alexander and Weissenborn, Dirk and Zhai, Xiaohua and Unterthiner, Thomas and Dehghani, Mostafa and Minderer, Matthias and Heigold, Georg and Gelly, Sylvain and Uszkoreit, Jakob and Houlsby, Neil},
  year = {2021},
  booktitle = {ICLR}
}

@inproceedings{misra2021endtoend,
  title = {An End-to-End Transformer Model for {3D} Object Detection},
  author = {Misra, Ishan and Girdhar, Rohit and Joulin, Armand},
  year = {2021},
  booktitle = {ICCV}
}

@inproceedings{qi2017pointnet,
  title = {{PointNet++}: Deep Hierarchical Feature Learning on Point Sets in a Metric Space},
  author = {Qi, Charles Ruizhongtai and Yi, Li and Su, Hao and Guibas, Leonidas J.},
  year = {2017},
  booktitle = {NeurIPS}
}

@inproceedings{yu2022pointbert,
  title = {{Point-BERT}: Pre-Training {3D} Point Cloud Transformers with Masked Point Modeling},
  author = {Yu, Xumin and Tang, Lulu and Rao, Yongming and Huang, Tiejun and Zhou, Jie and Lu, Jiwen},
  year = {2022},
  booktitle = {CVPR}
}

@inproceedings{zhou2024uni3d,
  title = {{Uni3D}: Exploring Unified {3D} Representation at Scale},
  author = {Zhou, Junsheng and Wang, Jinsheng and Ma, Baorui and Liu, Yu-Shen and Huang, Tiejun and Wang, Xinlong},
  year = {2024},
  booktitle = {ICLR}
}

@misc{hong2026r3d,
  title = {{R3D}: Revisiting {3D} Policy Learning},
  author = {Hong, Zhengdong and Wu, Shenrui and Cui, Haozhe and Zhao, Boyi and Ji, Ran and He, Yiyang and Zhang, Hangxing and Ke, Zundong and Wang, Jun and Zhang, Guofeng and Gu, Jiayuan},
  year = {2026},
  eprint = {2604.15281},
  archivePrefix = {arXiv}
}

@inproceedings{qi2017pointnetcvpr,
  title = {{PointNet}: Deep Learning on Point Sets for {3D} Classification and Segmentation},
  author = {Qi, Charles Ruizhongtai and Su, Hao and Mo, Kaichun and Guibas, Leonidas J.},
  year = {2017},
  booktitle = {CVPR}
}

@article{johnson1984extensions,
  title     = {Extensions of {Lipschitz} mappings into a {Hilbert} space},
  author    = {Johnson, William B. and Lindenstrauss, Joram},
  journal   = {Contemporary Mathematics},
  volume    = {26},
  pages     = {189--206},
  year      = {1984}
}

@inproceedings{frans2025one,
  title={One step diffusion via shortcut models},
  author={Frans, Kevin and Hafner, Danijar and Levine, Sergey and Abbeel, Pieter},
  booktitle={International Conference on Learning Representations},
  volume={2025},
  pages={34668--34684},
  year={2025}
}

@inproceedings{park2025ogbench,
  title     = {{OGBench}: Benchmarking Offline Goal-Conditioned {RL}},
  author    = {Park, Seohong and Frans, Kevin and Eysenbach, Benjamin and Levine, Sergey},
  booktitle = {International Conference on Learning Representations (ICLR)},
  year      = {2025}
}

@inproceedings{chi2023diffusionpolicy,
  title     = {Diffusion Policy: Visuomotor Policy Learning via Action Diffusion},
  author    = {Chi, Cheng and Feng, Siyuan and Du, Yilun and Xu, Zhenjia and
               Cousineau, Eric and Burchfiel, Benjamin and Song, Shuran},
  booktitle = {Robotics: Science and Systems (RSS)},
  year      = {2023}
}

@inproceedings{todorov2012mujoco,
  title     = {{MuJoCo}: A Physics Engine for Model-Based Control},
  author    = {Todorov, Emanuel and Erez, Tom and Tassa, Yuval},
  booktitle = {IEEE/RSJ International Conference on Intelligent Robots and Systems (IROS)},
  pages     = {5026--5033},
  year      = {2012}
}

@inproceedings{paszke2019pytorch,
  title = {{PyTorch}: An Imperative Style, High-Performance Deep Learning Library},
  author = {Paszke, Adam and Gross, Sam and Massa, Francisco and Lerer, Adam and Bradbury, James and Chanan, Gregory and Killeen, Trevor and Lin, Zeming and Gimelshein, Natalia and Antiga, Luca and Desmaison, Alban and K{\"o}pf, Andreas and Yang, Edward and DeVito, Zachary and Raison, Martin and Tejani, Alykhan and Chilamkurthy, Sasank and Steiner, Benoit and Fang, Lu and Bai, Junjie and Chintala, Soumith},
  year = {2019},
  booktitle = {NeurIPS}
}

@inproceedings{radford2021learning,
  title={Learning transferable visual models from natural language supervision},
  author={Radford, Alec and Kim, Jong Wook and Hallacy, Chris and Ramesh, Aditya and Goh, Gabriel and Agarwal, Sandhini and Sastry, Girish and Askell, Amanda and Mishkin, Pamela and Clark, Jack and others},
  booktitle={International conference on machine learning},
  pages={8748--8763},
  year={2021},
  organization={PmLR}
}

@inproceedings{kirillov2023segment,
  title={Segment anything},
  author={Kirillov, Alexander and Mintun, Eric and Ravi, Nikhila and Mao, Hanzi and Rolland, Chloe and Gustafson, Laura and Xiao, Tete and Whitehead, Spencer and Berg, Alexander C and Lo, Wan-Yen and others},
  booktitle={2023 IEEE/CVF international conference on computer vision (ICCV)},
  pages={3992--4003},
  year={2023},
  organization={IEEE}
}

@article{oquab2023dinov2,
  title={{DINOv2}: Learning Robust Visual Features without Supervision},
  author={Oquab, Maxime and Darcet, Timoth{\'e}e and Moutakanni, Th{\'e}o and Vo, Huy and Szafraniec, Marc and Khalidov, Vasil and Fernandez, Pierre and Haziza, Daniel and Massa, Francisco and El-Nouby, Alaaeldin and others},
  journal={arXiv preprint arXiv:2304.07193},
  year={2023}
}

@inproceedings{zeng2020transporter,
  title     = {Transporter Networks: Rearranging the Visual World for Robotic Manipulation},
  author    = {Zeng, Andy and Florence, Pete and Tompson, Jonathan and Welker, Stefan and
               Chien, Jonathan and Attarian, Maria and Armstrong, Travis and Krasin, Ivan and
               Duong, Dan and Sindhwani, Vikas and Lee, Johnny},
  booktitle = {Conference on Robot Learning (CoRL)},
  year      = {2020}
}

@inproceedings{sundermeyer2021contactgraspnet,
  title     = {{Contact-GraspNet}: Efficient {6-DoF} Grasp Generation in Cluttered Scenes},
  author    = {Sundermeyer, Martin and Mousavian, Arsalan and Triebel, Rudolph and Fox, Dieter},
  booktitle = {IEEE International Conference on Robotics and Automation (ICRA)},
  year      = {2021}
}

@inproceedings{nair2018visual,
  title     = {Visual Reinforcement Learning with Imagined Goals},
  author    = {Nair, Ashvin V. and Pong, Vitchyr and Dalal, Murtaza and Bahl, Shikhar and
               Lin, Steven and Levine, Sergey},
  booktitle = {NeurIPS},
  year      = {2018}
}

@inproceedings{du2023unipi,
  title     = {Learning Universal Policies via Text-Guided Video Generation},
  author    = {Du, Yilun and Yang, Sherry and Dai, Bo and Dai, Hanjun and Nachum, Ofir and
               Tenenbaum, Joshua B. and Schuurmans, Dale and Abbeel, Pieter},
  booktitle = {NeurIPS},
  year      = {2023}
}

@inproceedings{black2024susie,
  title     = {Zero-Shot Robotic Manipulation with Pretrained Image-Editing Diffusion Models},
  author    = {Black, Kevin and Nakamoto, Mitsuhiko and Atreya, Pranav and Walke, Homer and
               Finn, Chelsea and Kumar, Aviral and Levine, Sergey},
  booktitle = {ICLR},
  year      = {2024}
}

@article{thalhammer2024challenges,
  title   = {Challenges for Monocular {6-D} Object Pose Estimation in Robotics},
  author  = {Thalhammer, Stefan and Bauer, Dominik and H{\"o}nig, Peter and Weibel, Jean-Baptiste and
             Garc{\'\i}a-Rodr{\'\i}guez, Jos{\'e} and Vincze, Markus},
  journal = {IEEE Transactions on Robotics},
  volume  = {40},
  year    = {2024}
}

@article{newbury2023grasp,
  title   = {Deep Learning Approaches to Grasp Synthesis: A Review},
  author  = {Newbury, Rhys and Gu, Morris and Chumbley, Lachlan and Mousavian, Arsalan and
             Eppner, Clemens and Leitner, J{\"u}rgen and Bohg, Jeannette and Morales, Antonio and
             Asfour, Tamim and Kragic, Danica and Fox, Dieter and Cosgun, Akansel},
  journal = {IEEE Transactions on Robotics},
  volume  = {39},
  number  = {5},
  year    = {2023}
}

@article{demsar2006statistical,
  title   = {Statistical Comparisons of Classifiers over Multiple Data Sets},
  author  = {Dem{\v{s}}ar, Janez},
  journal = {Journal of Machine Learning Research},
  volume  = {7},
  pages   = {1--30},
  year    = {2006}
}

@article{wilcoxon1945individual,
  title   = {Individual Comparisons by Ranking Methods},
  author  = {Wilcoxon, Frank},
  journal = {Biometrics Bulletin},
  volume  = {1},
  number  = {6},
  pages   = {80--83},
  year    = {1945}
}

@article{holm1979simple,
  title   = {A Simple Sequentially Rejective Multiple Test Procedure},
  author  = {Holm, Sture},
  journal = {Scandinavian Journal of Statistics},
  volume  = {6},
  number  = {2},
  pages   = {65--70},
  year    = {1979}
}

@article{arnaud2025locate,
  title={Locate 3d: Real-world object localization via self-supervised learning in 3d},
  author={Arnaud, Sergio and McVay, Paul and Martin, Ada and Majumdar, Arjun and Jatavallabhula, Krishna Murthy and Thomas, Phillip and Partsey, Ruslan and Dugas, Daniel and Gejji, Abha and Sax, Alexander and others},
  journal={arXiv preprint arXiv:2504.14151},
  year={2025}
}

@inproceedings{saito2025point,
  title={Point-jepa: A joint embedding predictive architecture for self-supervised learning on point cloud},
  author={Saito, Ayumu and Kudeshia, Prachi and Poovvancheri, Jiju},
  booktitle={2025 IEEE/CVF Winter Conference on Applications of Computer Vision (WACV)},
  pages={7348--7357},
  year={2025},
  organization={IEEE}
}

@article{ye2026world,
  title={World action models are zero-shot policies},
  author={Ye, Seonghyeon and Ge, Yunhao and Zheng, Kaiyuan and Gao, Shenyuan and Yu, Sihyun and Kurian, George and Indupuru, Suneel and Tan, You Liang and Zhu, Chuning and Xiang, Jiannan and others},
  journal={arXiv preprint arXiv:2602.15922},
  year={2026}
}

@article{li2026causal,
  title={Causal world modeling for robot control},
  author={Li, Lin and Zhang, Qihang and Luo, Yiming and Yang, Shuai and Wang, Ruilin and Han, Fei and Yu, Mingrui and Gao, Zelin and Xue, Nan and Zhu, Xing and others},
  journal={arXiv preprint arXiv:2601.21998},
  year={2026}
}

@inproceedings{zhou2019continuity,
  title={On the continuity of rotation representations in neural networks},
  author={Zhou, Yi and Barnes, Connelly and Lu, Jingwan and Yang, Jimei and Li, Hao},
  booktitle={2019 IEEE/CVF Conference on Computer Vision and Pattern Recognition (CVPR)},
  pages={5738--5746},
  year={2019},
  organization={IEEE}
}

@inproceedings{fang2024rh20t,
  title={Rh20t: A comprehensive robotic dataset for learning diverse skills in one-shot},
  author={Fang, Hao-Shu and Fang, Hongjie and Tang, Zhenyu and Liu, Jirong and Wang, Chenxi and Wang, Junbo and Zhu, Haoyi and Lu, Cewu},
  booktitle={2024 IEEE International Conference on Robotics and Automation (ICRA)},
  pages={653--660},
  year={2024},
  organization={IEEE}
}

@inproceedings{dehghani2023scaling,
  title     = {Scaling Vision Transformers to 22 Billion Parameters},
  author    = {Dehghani, Mostafa and Djolonga, Josip and Mustafa, Basil and
               Padlewski, Piotr and Heek, Jonathan and others},
  booktitle = {Proceedings of the 40th International Conference on Machine Learning},
  series    = {PMLR},
  volume    = {202},
  pages     = {7480--7512},
  year      = {2023}
}

@inproceedings{wortsman2024small,
  title     = {Small-scale Proxies for Large-scale Transformer Training Instabilities},
  author    = {Wortsman, Mitchell and Liu, Peter J. and Xiao, Lechao and Everett, Katie
               and Alemi, Alexander and Adlam, Ben and Co-Reyes, John D. and Gur, Izzeddin
               and Kumar, Abhishek and Novak, Roman and Pennington, Jeffrey
               and Sohl-Dickstein, Jascha and Xu, Kelvin and Lee, Jaehoon
               and Gilmer, Justin and Kornblith, Simon},
  booktitle = {International Conference on Learning Representations (ICLR)},
  year      = {2024}
}

@inproceedings{henry2020qknorm,
  title     = {Query-Key Normalization for Transformers},
  author    = {Henry, Alex and Dachapally, Prudhvi Raj and Pawar, Shubham Shantaram
               and Chen, Yuxuan},
  booktitle = {Findings of the Association for Computational Linguistics: EMNLP 2020},
  pages     = {4246--4253},
  year      = {2020},
  doi       = {10.18653/v1/2020.findings-emnlp.379}
}
\bibliographystyle{iclr2027_conference}
\newpage

\appendix
\section*{Appendix}
This appendix collects the material that supports the main text.
Appendix~\ref{app:impl} lists the architectures, optimization constants, and
compute behind every model. Appendix~\ref{app:goal-ambiguity} formalizes why
the target-to-latent module conditions on the current latent, and
Appendix~\ref{app:pc-detail} details how the four benchmark datasets are
re-sensed as point clouds. Appendix~\ref{app:experiments} probes the latents
for physical state, perturbs the point clouds at test time, and visualizes
where the encoders attend. Appendix~\ref{app:stats} backs the main
comparisons with paired significance tests, and Appendix~\ref{app:impact}
discusses broader impacts.

\section{Implementation details}
\label{app:impl}
This section records the architecture and optimization constants behind
Section~\ref{sec:models}. The released configurations remain authoritative.
All world models are trained on two A100 GPUs.

\subsection{Point-cloud encoder (\plewm{}, \pdjepa{})}
\label{app:point-encoder}
The shared encoder (Figure~\ref{fig:arch_pointvit}) consumes geometry only:
the per-point input features are the sensor-frame coordinates, with no
intensity or color channel. Each cloud is reduced to 256 farthest-point-sampled
centres, and a ball query of per-environment metric radius (0.15\,m Two-Room,
0.18\,m Push-T, 0.04\,m Reacher) draws up to 32 points per centre uniformly at
random \citep{qi2017pointnet}. Coordinates are first normalized by the affine transformation $\vp\mapsto(\vp-\vmu)/s$ with a
per-environment centre $\vmu\in\R^3$ and isotropic scale $s$ measured
once on the dataset, which brings the workspace to approximately $[-1,1]^3$
while preserving absolute object positions. The centre-relative offsets of
each group then pass through a two-stage mini-PointNet
\citep{qi2017pointnetcvpr,yu2022pointbert}
with hidden widths $(128, 256)$, yielding one 192-dimensional token to which a
log-spaced sinusoidal embedding of the group centre \citep{misra2021endtoend}
is added. The tokens, prepended by a learned \textsc{cls} token, feed a
LayerNorm-only \citep{hong2026r3d} ViT-Tiny trunk (12 layers, 3 heads, width
192, ${\approx}5.5$M parameters) \citep{dosovitskiy2021image}, whose
\textsc{cls} output is the latent $\vz_t$. OGB-Cube uses 512 tokens with radius
0.03\,m instead. The denser sampling helped only there, since the other
scenes are dominated by the ground plane.

The predictor is LeWorldModel's causal action-conditioned transformer
\citep{maes2026leworldmodel}: 6 blocks, 16 heads of dimension 64, MLP width
2048 at the 192-dimensional latent width, with a context of $\ell=3$ past latents.
Latents pass a BatchNorm MLP projector (hidden 2048), as in the image model.
The world models step at frameskip 5, the five intervening environment
actions concatenated and embedded by a shared linear action encoder. Both
end-to-end models train for 10 epochs with AdamW (learning rate
$5\times10^{-5}$, weight decay $10^{-3}$) at effective batch size 128, bf16
precision, and gradient clipping 1.0. The SIGReg weight is $\lambda=0.09$ (17 knots,
1024 projections) \citep{balestriero2025lejepa,maes2026leworldmodel}. The
action-reconstruction weight is $\beta=10$ at horizon $N{=}5$
\citep{zhang2026deltajepa}, for which \pdjepa{} trains on 6-frame windows while
planning with the same 3-latent context. On OGB-Cube both models follow a
50-epoch cosine schedule, with the reported checkpoints still taken after 10
epochs.

\subsection{\utoniawm{}}
\label{app:utonia-wm}
Completing Section~\ref{sec:models}: point clouds are expressed in a canonical frame
derived from the fitted ground plane, rescaled by Utonia's own tabletop factor
of 4.0, and voxel-downsampled at a grid size of 0.01, keeping one point per
occupied voxel, before entering the frozen ${\approx}137$M-parameter PTv3
backbone \citep{wu2023point,zhang2026utonia}.
Per grid cell, the backbone's coarsest-stage 576-dimensional super-point
features are mean- and max-pooled (1152 dimensions) and compressed to the
192-dimensional token width by the fixed random orthogonal projection of
Section~\ref{sec:models}, a Gaussian matrix drawn once from a fixed seed and
orthonormalized via QR. The grids are $8\times8\times4=256$ tokens
(OGB-Cube, Reacher), $12\times11\times2=264$ (Push-T), and
$10\times9\times3=270$ (Two-Room), spanning a
per-environment workspace box. The predictor is DINO-WM's frame-block-causal
token transformer \citep{zhou2025dinowm} at the same depth and widths as above,
with the action embedding tiled over a frame's tokens and concatenated onto
each. Only the predictor and action encoder train, against precomputed frozen
features, for 100 epochs with AdamW (learning rate $10^{-3}$, weight decay
$10^{-3}$) at effective batch size 512.

\paragraph{Utonia-WM ablation.}                                                                                 
Table~\ref{tab:utonia_ablation} ablates the three main design choices of the
\utoniawm{} encoder. The \emph{voxel resolution} is the $xy$ cell size of the
grid that pools Utonia's super-point features into tokens. The coarse, base,
and fine grids have $4{\times}4{\times}4$, $8{\times}8{\times}4$, and
$12{\times}12{\times}4$ tokens on OGB-Cube and $6{\times}6{\times}2$,
$12{\times}11{\times}2$, and $18{\times}17{\times}2$ tokens on Push-T. The
\emph{feature hierarchy} determines which PTv3 stages are pooled into each
token, which is either the coarsest stage alone or all five stages
concatenated (\emph{Hypercolumn} in Table \ref{tab:utonia_ablation}. The \emph{projection} reduces each token to
192 dimensions with either a seeded random orthogonal matrix or a PCA fitted
on 2\,000 training frames. We vary each choice separately and additionally
combine hypercolumn features with PCA.

Every configuration is trained on a 50-epoch schedule. The retrained base
recipe closely matches the 100-epoch results of Table~\ref{tab:main} (70.8
vs.\ 71.2 on OGB-Cube, 47.6 vs.\ 46.6 on Push-T), so the shorter schedule
does not disadvantage it. No configuration is significantly better than the
base recipe on either environment. OGB-Cube is insensitive to all three
choices. Push-T degrades when the grid is coarsened, because cells of
${\approx}83{\times}71$\,cm exceed the size of the T-block. It also degrades
when hypercolumn features are combined with PCA.

With the finer grid, the sequences over three frames grow to 1,728 and 1,836 tokens. At this length, the predictor collapses to predicting the mean token. In the collapsed runs, the query and key weights grow so large that the attention logits can reach the thousands, whereas they stay below one hundred in the base configuration. Logit growth of this kind is known to saturate the softmax and destabilize training \citep{dehghani2023scaling, wortsman2024small}. Query-key normalization \citep{henry2020qknorm,dehghani2023scaling} could resolve this, but we do not adopt it because it adds runtime and memory overhead in our implementation. In any case, the finer grid is not a candidate for Table~\ref{tab:main}, since its extra tokens would give \utoniawm{} an unfair advantage over DINO-WM and would further increase planning time (Table~\ref{tab:timing}). We therefore use the base configuration, which also has the benefit of requiring no fitted projection.

\begin{table}[t]
  \centering
  \small
  \caption{\textbf{Utonia-WM encoder ablation.} Planning success rate (\%),
    mean$_{\pm\text{sd}}$ over 10 seeds of 50 episodes. Each row changes one
    factor of the base recipe used in Table~\ref{tab:main}, and \checkmark/\ding{55}
    mark the variant used in each configuration. \emph{Voxel resolution}
    scales the $xy$ cell size of the token grid inside the same workspace box.
    \emph{Feature hierarchy} is the set of PTv3 stages pooled into a token
    (coarsest stage, or all five stages as a hypercolumn). \emph{Projection}
    is either seeded random orthogonal or PCA fitted on
    2\,000 training frames). All cells train on a 50-epoch schedule.}
  \label{tab:utonia_ablation}
  \newcommand{\yes}{\checkmark}
  \newcommand{\no}{\ding{55}}
  \newcommand{\sig}{$^{\dagger}$}
  \newcommand{\sigb}{$^{\ddagger}$}
  \newcommand{\hd}[2]{\shortstack{#1\\#2}}
  \setlength{\tabcolsep}{3pt}
  \begin{tabular*}{\linewidth}{@{\extracolsep{\fill}}l ccc cc cc ll@{}}
    \toprule
    & \multicolumn{3}{c}{\textbf{Voxel resolution}}
      & \multicolumn{2}{c}{\textbf{Feature hierarchy}}
      & \multicolumn{2}{c}{\textbf{Projection}}
      & \multicolumn{2}{c}{\textbf{Success (\%)}} \\
    \cmidrule(lr){2-4} \cmidrule(lr){5-6} \cmidrule(lr){7-8} \cmidrule(l){9-10}
    Configuration
      & \hd{Coarse}{($0.5\times$)} & \hd{Base}{($1\times$)} & \hd{Fine}{($1.5\times$)}
      & \hd{Coarsest}{stage} & \hd{Hyper-}{column}
      & \hd{Rand.}{orth.} & \hd{PCA}{\vphantom{(}}
      & OGB-Cube & Push-T \\
    \midrule
    Base recipe       & \no  & \yes & \no  & \yes & \no  & \yes & \no  &
      70.8\sd{4.8} & 47.6\sd{6.2} \\
    \addlinespace[3pt]
    PCA projection    & \no  & \yes & \no  & \yes & \no  & \no  & \yes &
      68.2\sd{6.5} & 44.2\sd{4.0} \\
    Hypercolumn       & \no  & \yes & \no  & \no  & \yes & \yes & \no  &
      70.4\sd{7.6} & 44.4\sd{6.2} \\
    Hypercolumn + PCA & \no  & \yes & \no  & \no  & \yes & \no  & \yes &
      73.4\sd{6.0} & 40.2\sd{3.2}\sigb \\
    \addlinespace[3pt]
    Coarser grid      & \yes & \no  & \no  & \yes & \no  & \yes & \no  &
      71.0\sd{7.6} & 26.0\sd{5.6}\sigb \\
    Finer grid        & \no  & \no  & \yes & \yes & \no  & \yes & \no  &
      \multicolumn{1}{c}{--\sig} & \multicolumn{1}{c@{}}{--\sig} \\
    \bottomrule
  \end{tabular*}
  \par\smallskip
  {\footnotesize $^{\dagger}$ Collapsed to mean prediction under the shared recipe (see text). $^{\ddagger}$Significantly below base (paired $t$-test over the
  same seeds, $p<0.01$); all other differences are within seed noise ($p>0.1$).}
\end{table}

\subsection{\voxwm{}}
\label{app:vox-wm}
\voxwm{} is a parameter-free geometric control for \utoniawm{} that ablates
the frozen foundation encoder at fixed tokenization: same canonical frame,
same token grid, same predictor and prediction-only loss, but each cell's
token holds plain statistics of the cell's points instead of pooled frozen
features. After the voxel downsampling above, a cell's points are expressed as
offsets from the cell centre, and the 25-dimensional token stacks an occupancy
bit (1), the point count as $\log(1{+}n)/\log(1{+}256)$ (1), the per-axis mean
(3), standard deviation (3), minimum (3), and maximum (3) of the offsets,
their three off-diagonal covariances (3), and the occupancy bits of a
$2\times2\times2$ subdivision of the cell (8). Empty cells are zero tokens,
which the occupancy bit disambiguates.
The 25 statistics are lifted to the
192-dimensional predictor width by a fixed random orthogonal projection,
mirroring the one above, so both models present an identical training-free
interface to the predictor and differ only in what fills each token.
The
encoding is deterministic and costs microseconds per cloud, so it runs live
without a feature cache. Training follows the \utoniawm{} protocol. 

\subsection{Planning}
\label{app:planning}
All methods plan with the receding-horizon model predictive control (MPC)
loop of the SWM framework \citep{maes2026stableworldmodel},
with one configuration in all four environments. At each planning call, the
policy encodes the goal observation once and conditions the predictor on the
same three-frame observation history used in training. The cross-entropy
method (CEM) then searches for an action sequence that minimizes the squared
distance between the predicted latent at the end of the horizon and the goal
latent (Section~\ref{sec:models}). Only the final predicted step enters the
cost. CEM maintains a diagonal Gaussian over standardized actions,
initialized with zero mean and unit variance. Each iteration samples 300
candidate sequences, always including the current mean, scores them in one
batched rollout of the predictor, and refits the mean and standard deviation
to the 30 lowest-cost candidates. After 30 iterations the mean of the
distribution is returned as the plan. The planning horizon is 5 steps, and
each planned action is repeated for 5 environment steps (frameskip 5), so
one plan covers 25 environment steps. The agent executes the full plan and
then replans from the new observation without warm-starting from the
previous plan. The goal is the state 25 steps ahead of the start along a
trajectory of the training dataset, and success is reaching it within a
budget of 50 environment steps, so at most two plans are computed per
episode. Every model is evaluated on the same 50 start states under each of
the same ten seeds, so the image and point-cloud comparisons of
Table~\ref{tab:main} are paired under one protocol.

\subsection{Target-to-latent module}
\label{app:target-to-latent}
Both instantiations of $m_\xi$ (Section~\ref{sec:models}) share one
conditioning interface. The target coordinates, namely the cube position
(3-d), the Two-Room agent position (3-d), the Reacher joint positions (6-d), or the Push-T agent and block
positions together with the block's heading vector (9-d), are expressed in the
sensor frame, normalized by the same workspace map $\vp\mapsto(\vp-\vmu)/s$ as the
encoder inputs, and lifted by log-spaced sinusoidal Fourier features (32
bands) \citep{misra2021endtoend}, which make
sub-centimetre offsets linearly separable. The cube target carries no
orientation channel, for the symmetry reason of Appendix~\ref{app:probing}.
The \emph{MLP} head is a residual MLP (width 1024, hidden 2048, 4 blocks)
regressing the normalized goal latent. The heading channels and, when used,
$\vz_t$ enter through a linear embedding added to the Fourier code. 

The \emph{shortcut} head \citep{frans2025one} is an adaptive-LayerNorm-modulated
residual MLP of the same widths over the normalized latent space, trained with
flow matching plus the self-consistency loss against an EMA copy (decay
0.999). Here, the heading channels
and $\vz_t$ join the conditioning vector that modulates every block. Training
pairs are frames of one trajectory separated by $\pm5$ to $\pm25$ environment
steps, with $\vz_t$ the latent of the earlier frame and the target read from
the later. At evaluation $\vz_t$ is re-encoded at every replan. 

All heads train for
60 epochs with AdamW (learning rate $3\times10^{-4}$, weight decay $10^{-4}$)
at batch size 4096 on precomputed frozen latents, with 5\,mm Gaussian jitter
on the position channels, without which the Fourier code is sharp enough to
memorize individual frames.
During evaluation, the shortcut model generates a latent sample in a single inference step. Its inference time is comparable to that of the MLP baseline, which directly predicts the latent in one forward pass.

\subsection{Compute \& Runtime}
\label{app:compute}
Every world model trains on two A100 GPUs of an internal cluster. The time budget for end-to-end training runs was set to 72 hours. One evaluation sweep of 10 seeds $\times$ 50 episodes costs a few GPU-hours per model and environment, considerably more for \utoniawm{}. 

Table~\ref{tab:timing} reports the inference and planning cost of the studied world models. All timings are measured on one NVIDIA RTX~4090 with
an Intel Xeon W-2235 host CPU. For each checkpoint, we time 300 encoder
and 300 predictor calls, and a total of 30 plans on the same inputs across all environments. The table lists the mean and standard deviation of the runs per configuration.
The encoder time measures mapping one frame to its latent. The
point-cloud encoders of P-LeWM and PD-JEPA tokenize the cloud into 256
farthest-point groups. We also evaluate a variant using 512
groups. The predictor step is one autoregressive step for all 300 CEM candidates with the full three-frame
history window. Finally, the plan time is one complete CEM solve for one episode with
300 candidates, 30 iterations, and a horizon of 5, which is the planner used
in all evaluations.

\plewm{}, \pdjepa{}, and their image counterparts need about 3 ms per predictor
step and plan in about 0.5 s. Doubling the number of point tokens from 256
to 512 raises the encoder time from roughly 10.5 to 14 ms but leaves the
predictor and planning cost unchanged. DINO-WM, \utoniawm{}, and Vox-WM
need about 565 ms per predictor step and more than 60 s per plan. Their planning cost is dominated by the predictor, so even the 90 ms Utonia encoder contributes little to the total. The larger standard deviations of \utoniawm{} and Vox-WM arise because their token grid size differs between environments.

\begin{table}[t]
    \centering
    \caption{Inference and planning cost of the compared world models on one
      NVIDIA RTX~4090. Mean $\pm$ standard deviation over repeated calls,
      pooled over the checkpoints of the four environments. The encoder time
      covers one frame, the predictor step covers one autoregressive step for
      all 300 CEM candidates, and the plan time covers one complete CEM solve
      (300 candidates, 30 iterations, horizon 5).}
    \label{tab:timing}
    \begin{tabular}{llrrr}
      \toprule
      Model & Input & Encoder [ms] & Predictor step [ms] & Plan [s] \\
      \midrule
      P-LeWM (256 tokens) & point cloud & 10.5 $\pm$ 0.2 & 3.0 $\pm$ 0.1 & 0.51 $\pm$ 0.01 \\
      P-LeWM (512 tokens) & point cloud & 14.2 $\pm$ 0.2 & 3.0 $\pm$ 0.1 & 0.53 $\pm$ 0.02 \\
      PD-JEPA (256 tokens) & point cloud & 10.6 $\pm$ 0.3 & 3.1 $\pm$ 0.1 & 0.52 $\pm$ 0.01 \\
      PD-JEPA (512 tokens) & point cloud & 14.1 $\pm$ 0.4 & 2.9 $\pm$ 0.1 & 0.52 $\pm$ 0.01 \\
      Image LeWM & RGB & 4.1 $\pm$ 0.2 & 2.9 $\pm$ 0.1 & 0.48 $\pm$ 0.02 \\ 
      Image Delta-JEPA & RGB & 4.1 $\pm$ 0.2 & 3.0 $\pm$ 0.1 & 0.51 $\pm$ 0.02 \\
      \midrule
      DINO-WM & RGB & 5.0 $\pm$ 0.2 & 565 $\pm$ 3 & 61.6 $\pm$ 0.1 \\
      Utonia-WM & point cloud & 90.3 $\pm$ 2.9 & 563 $\pm$ 22 & 63.9 $\pm$ 2.6 \\
      Vox-WM & point cloud & 1.8 $\pm$ 0.1 & 564 $\pm$ 23 & 64.0 $\pm$ 2.7 \\
      \bottomrule 
    \end{tabular}
  \end{table}

\subsection{Fidelity of the image baselines}
\label{app:baseline-fidelity}

Table~\ref{tab:baseline-fidelity} places the image baselines of
Table~\ref{tab:main} next to the success rates their authors report.
Retrained Delta-JEPA matches or exceeds its published rates on all four
benchmarks, by $+4.7$ points on Reacher and $+5.1$ points on Push-T. Point-Delta-JEPA in Table~\ref{tab:main} is
therefore measured against a baseline at full strength. Retrained DINO-WM
is within two points of its published rate on Reacher and Push-T and
$7.4$ points below on OGB-Cube. The released LeWM checkpoint is within four
points of its published rate on Two-Room, Reacher and OGB-Cube and $9.6$
points below on Push-T. An independent retraining of LeWM by
\citet{zhang2026deltajepa} lands at $84.5$ on Push-T and $64.1$ on
OGB-Cube, in line with our $86.4$ and $69.8$ and below the published
$96$ and $74$.

These offsets are consistent with the evaluation protocols.
\citet{maes2026leworldmodel} report a single evaluation seed of 50
trajectories, \citet{zhang2026deltajepa} report three seeds of 500
trajectories each sampled by them, and we report ten seeds of 50 episodes
on fixed start--goal pairs shared by every model. On OGB-Cube our seed
standard deviations range from $5.9$ to $9.3$ points, so a single seed of
50 episodes can sit a full offset above or below the ten-seed mean, and
both independent evaluations of LeWM on that environment land below the
published rate. Two further differences apply. \citet{maes2026leworldmodel}
run 10 CEM iterations outside Push-T, whereas we run 30 everywhere. On
Two-Room they sample the goal 100 steps ahead with a 150-step budget,
whereas we keep the 25-step goal and 50-step budget of the other
environments, which is also why our Two-Room random baseline is $25\%$
rather than $0\%$. The comparisons in Table~\ref{tab:main} are paired
within one protocol and are unaffected by these offsets.

\begin{table}[h]
\centering
\caption{\textbf{Image baselines versus their published success rates (\%).}
Ours are the Table~\ref{tab:main} values under our protocol. Published
DINO-WM excludes proprioception, as in our runs. Two-Room protocols differ
(see text).}
\label{tab:baseline-fidelity}
\small
\begin{tabular}{llcccc}
\toprule
Model & Source & Two-Room & Reacher & Push-T & OGB-Cube \\
\midrule
Delta-JEPA & published \citep[Tab.~1]{zhang2026deltajepa} & 100.0 & 81.3 & 89.1 & 79.3 \\
           & ours (retrained)                               & 99.8  & 86.0 & 94.2 & 80.2 \\
\midrule
DINO-WM    & published \citep[Fig.~6]{maes2026leworldmodel} & 100   & 79   & 74   & 86   \\
           & ours (retrained)                                & 99.8  & 77.0 & 76.0 & 78.6 \\
\midrule
LeWM       & published \citep[Fig.~6]{maes2026leworldmodel}  & 87    & 86   & 96   & 74   \\
           & retrained by \citet[Tab.~1]{zhang2026deltajepa} & 74.9  & 79.9 & 84.5 & 64.1 \\
           & ours (released checkpoint)                      & 85.6  & 82.0 & 86.4 & 69.8 \\
\bottomrule
\end{tabular}
\end{table}

\subsection{Licenses}
\label{app:licenses}
All existing assets are used under their published licenses: the
stable-worldmodels suite (0.1.1) and its released LeWM checkpoints
\citep[][MIT]{maes2026stableworldmodel}, OGBench 1.2.1 \citep[][MIT]{park2025ogbench}, the Push-T environment \citep[][MIT]{chi2023diffusionpolicy}, MuJoCo
3.10.0 \citep[][Apache 2.0]{todorov2012mujoco}, the pretrained Utonia backbone
\citep[][CC-BY-NC-4.0]{zhang2026utonia}, and PyTorch 2.7.0
\citep[][BSD-3-Clause]{paszke2019pytorch}.

\section{The unconditioned target-to-latent objective}
\label{app:goal-ambiguity}

This section makes precise the claim of Section~\ref{sec:models} that, without
the current latent, the regression target of \eqref{eq:goal} need not be
the latent of any reachable observation. Write
$\gZ_f=\{f_\theta(o):o\in\operatorname{supp}\gD\}$ for the set
of latents the frozen encoder assigns to observations in the support of the
data distribution $\gD$, that is, to observations that actually occur
in the data, and let $(\vz_t, \vc_{t+k}, \vz_{t+k})$ be distributed as in
\eqref{eq:goal}. The unconditioned variant minimizes
\begin{equation}
\label{eq:goal-uncond}
\Ls_{\mathrm{goal}}^{\emptyset}(\xi)=\E\bigl[\lVert m_\xi(\vc_{t+k})-\vz_{t+k}\rVert_2^2\bigr],
\end{equation}
and over unconstrained functions $m:\gC\to\gZ$ this mean-squared error
is minimized pointwise by the conditional mean
\begin{equation}
\label{eq:goal-mean}
m^{\ast}(\vc)=\E\bigl[\vz_{t+k}\,\big|\,\vc_{t+k}=\vc\bigr],
\end{equation}
the barycenter of the latents of \emph{all} data scenes whose annotation
equals $\vc$, covering every configuration of the scene content the target
leaves free, on OGB-Cube every arm pose. $\gZ_f$ is the image of the
low-dimensional scene configuration manifold under the continuous encoder
$f_\theta$, hence itself a low-dimensional subset of $\R^{192}$, and
no term of the training objective encourages it to be convex. SIGReg in
particular shapes one-dimensional projections of the latent distribution
rather than the geometry of its support. A barycenter of distinct points of a
non-convex set need not lie in the set, and $m^{\ast}(\vc)$ can therefore be a
latent that no observation in the data distribution produces. A planner minimizing $\lVert\hat{\vz}_{t+H}-m^{\ast}(\vc)\rVert_2^2$ is
then steered toward a goal latent it may be unable to attain.

Conditioning changes the average, not the argument: with $\vz_t$ as input the
minimizer becomes
\begin{equation}
\label{eq:goal-cond-mean}
m^{\ast}(\vc,\vz_t)=\E\bigl[\vz_{t+k}\,\big|\,\vc_{t+k}=\vc,\,\vz_t\bigr],
\end{equation}
an average over futures of the current scene only. Scene content that $\vz_t$ determines and
that the dynamics leave unchanged over the offset window contributes no
variance to this average, so the residual spread stems only from the moving
entities the target leaves unconstrained. Whether that residue matters is an
empirical question, and Table~\ref{tab:target2latent} answers it per
environment: on Push-T and Two-Room the target constrains all moving entities
and the unconditioned variant already matches the goal-cloud reference, while
on OGB-Cube the arm remains free at the goal even given $\vz_t$, and dropping
$\vz_t$ costs up to 10 points. The two heads of Section~\ref{sec:models} also
treat the average differently: the MLP fits the conditional mean of
\eqref{eq:goal-mean} directly, while the shortcut model samples from the
conditional distribution and so returns one consistent latent rather than a
barycenter, although Table~\ref{tab:target2latent} separates the two heads
only within seed noise.

\paragraph{Why not synthesize the goal scan?}
Since we control the mirror geometry, one could place the task-relevant
entity at the specified target, cast a synthetic scan of that state, and
encode it, with no learned module. The construction founders on the same
ambiguity as the unconditioned regression: a scene must be fully configured
before a scan can be cast, the target constrains only the task-relevant
coordinates, so the synthesizer has to invent the remaining content, on
OGB-Cube an arm pose, and an invented configuration need not be reachable or
consistent with the present scene. The conditioned module resolves this from
$\vz_t$ without reconstructing geometry, and in deployment the mismatch grows,
since no perfect mirror of the scene exists to render from. The module itself
is also modality-agnostic: given the same target annotations it could be
trained on image latents. What geometry specifically contributes is that the
target already lives in the metric space of the observation, so no
cross-modal correspondence must be learned.

\section{Point-cloud sensing and dataset replication: details}
\label{app:pc-detail}

\begin{table}[t]
  \centering
  \caption{\textbf{The four re-sensed datasets.} \emph{Ground share} and \emph{moving
  returns} are the fractions of returns on the ground plane and on non-static
  entities. Both are constant within an environment.}
  \label{tab:pc-datasets-full}
  \begin{tabular}{lcccc}
    \toprule
     & OGB-Cube & Two-Room & Push-T & Reacher \\
    \midrule
    Sensor pose        & scene camera & derived oblique & derived oblique & derived oblique \\
    Ground share       & $\sim$79\% & $\sim$80\% & $\sim$98\% & $\sim$97\% \\
    Moving returns     & $\sim$15.0\% & $\sim$0.3\% & $\sim$0.5\% & $\sim$2.0\% \\
    Episodes           & 10{,}000 & 10{,}000 & 18{,}685 & 10{,}000 \\
    Frames             & 2{,}010{,}000 & 920{,}809 & 2{,}336{,}736 & 2{,}010{,}000 \\
    Episode length     & 201 & 31--101 & 49--246 & 201 \\
    \bottomrule
  \end{tabular}
\end{table}

Exact sensor parameters, mirror geometry and scene constants are given by the
released configurations. This section records only the choices behind them.

\paragraph{Ray budget.}
The tokenizer reduces every cloud to 256 farthest-point-sampling centres, so the
ray grid fixes the sampling density of the observation rather than the sequence
length seen by the model. We use $100\times100$ as the coarsest grid at which
the smallest task-relevant entity, the Push-T pusher or the Two-Room agent,
still returns on the order of a hundred points.

\paragraph{Viewpoint.}
Oblique viewpoints are derived rather than hand-placed: given bounding-sphere
radius $R$ the camera sits at $d = 2.5R$ along the $45^\circ$ azimuth and
elevation bearing, with vertical field of view $2\arcsin(R/d)$, so no pose is
tuned per environment. The OGBench cube is the exception: its scene already
contains the oblique camera its released images were rendered from, and the
sensor is mounted on it, so that its scan and the image baseline's
observation share a viewpoint.

\paragraph{Mirror scenes.}
Entity heights in the two mirrored environments are ours to choose rather than
inherited from the benchmark, and were fixed once per environment: large enough
that an entity is separable from the floor in an oblique scan, small enough that
self-occlusion does not dominate. Reacher3D is native, but the geometry of the
released environment did not match the dataset it shipped with. We recovered the
two link lengths by least-squares fitting to the recorded joint angles and
fingertip positions, after which forward kinematics reproduces the recorded
fingertip.

\paragraph{Goal handling.}
The goal marker is excluded from every scan, by an alpha threshold in the cube
and by omission from the mirror geometry elsewhere, since a goal specifies a
task rather than a surface a range sensor would return, and retaining it would
permit a policy to read the goal off the observation rather than infer it. The
goal reaches the planner as the stored cloud of the goal frame, indexed from the
same converted table by the same goal offset the image pipeline uses, so no
separate goal scan is cast.

\section{Experiments}
\label{app:experiments}

\subsection{Physical latent probing}
\label{app:probing}

\paragraph{Protocol.}
For each environment the frozen checkpoints of the two end-to-end world models
encode every tenth frame of its training table. The probed representation is
$\vz=f_\theta(o)$ after the projector, i.e.\ exactly the 192-dimensional latent
the CEM cost operates on. Two probes per target regress privileged simulator
state from $\vz$: a \emph{linear} probe, closed-form ridge regression on
standardized features with the regularization weight selected on a validation
split over a logarithmic grid, and an \emph{MLP} probe
($192\!\to\!512\!\to\!512\!\to\!d$, GELU, AdamW with cosine schedule,
early-stopped on the same validation split). Splits are drawn at
\emph{episode} granularity (72/8/20 train/val/test), so every test frame comes
from a trajectory no probe ever saw. We report test-set MSE in native units
and pooled $R^2$. All targets live in the sensor frame of the stored clouds,
and planar angles enter as unit heading vectors, $(\cos\gamma,\sin\gamma,0)$ for a planar angle $\gamma$
rotated into the sensor frame, so that no probe has to fit a wrap-around
discontinuity.

\paragraph{Results.}
Table~\ref{tab:probe-pusht} reports Push-T, analyzed in the main text.
Tables~\ref{tab:probe-reacher} and~\ref{tab:probe-cube} complement it on
Reacher and OGB-Cube. Reacher is saturated: the arm is fully visible from
the oblique viewpoint, and every property, positions and link headings
alike, is linearly decodable at $R^2\!\approx\!1$ from both models, which
is why it offers no signal for separating the objectives. On OGB-Cube the two
yaw rows require care: a cube is 24-fold rotationally symmetric, so its
orientation is not identifiable from geometry at all, and the raw yaw label is
not folded into a fundamental domain. The near-zero $R^2$ therefore
reflects an unidentifiable \emph{label}, not a representation failure. The
end-effector yaw is similarly flat for both models. The wrist subtends only a
handful of returns at the $100\times100$ ray budget, and neither objective
rewards retaining it. Joint velocity, finally, is not observable from a single
scan and can be decoded only through its correlation with configuration along
the data distribution. That \pdjepa{} alone recovers a substantial share of it
($R^2$ 0.63 vs.\ 0.10 with the MLP probe) is consistent with its objective,
which forces the latent \emph{displacement} to be decodable into the executed
actions.

% Push-T probing
\begin{table}[t]
\centering
\caption{\textbf{Physical latent probing results on Push-T.}
All targets are 3-dimensional: locations are coordinates in the sensor frame, and the block angle
is encoded as a heading vector in the sensor frame. Lower MSE and higher $R^2$ indicate better
representation quality.}
\label{tab:probe-pusht}
\begin{tabular}{llcccc}
\toprule
& & \multicolumn{2}{c}{\textbf{Linear}} & \multicolumn{2}{c}{\textbf{MLP}} \\
\cmidrule(lr){3-4} \cmidrule(lr){5-6}
\textbf{Property} & \textbf{Method} & \textbf{MSE} $\downarrow$ & $R^2$ $\uparrow$ & \textbf{MSE} $\downarrow$ & $R^2$ $\uparrow$ \\
\midrule
\multirow{2}{*}{Agent Location}
 & Point-LeWM & 0.008 & 0.955 & \textbf{0.001} & \textbf{0.997} \\
 & Point-Delta-JEPA & \textbf{0.007} & \textbf{0.957} & 0.002 & 0.989 \\
\midrule
\multirow{2}{*}{Block Location}
 & Point-LeWM & \textbf{0.004} & \textbf{0.958} & \textbf{0.000} & \textbf{0.998} \\
 & Point-Delta-JEPA & 0.005 & 0.946 & 0.001 & 0.992 \\
\midrule
\multirow{2}{*}{Block Angle}
 & Point-LeWM & \textbf{0.027} & \textbf{0.891} & \textbf{0.002} & \textbf{0.992} \\
 & Point-Delta-JEPA & 0.084 & 0.657 & 0.009 & 0.962 \\
\bottomrule
\end{tabular}
\end{table}

% Reacher probing
\begin{table}[t]
\centering
\caption{\textbf{Physical latent probing results on Reacher.}
Positions are 3D coordinates in the sensor frame, and joint angles are encoded as per-link unit
heading vectors in the sensor frame. Lower MSE and higher $R^2$ indicate better representation
quality.}
\label{tab:probe-reacher}
\begin{tabular}{llcccc}
\toprule
& & \multicolumn{2}{c}{\textbf{Linear}} & \multicolumn{2}{c}{\textbf{MLP}} \\
\cmidrule(lr){3-4} \cmidrule(lr){5-6}
\textbf{Property} & \textbf{Method} & \textbf{MSE} $\downarrow$ & $R^2$ $\uparrow$ & \textbf{MSE} $\downarrow$ & $R^2$ $\uparrow$ \\
\midrule
\multirow{2}{*}{Finger Position}
 & Point-LeWM & \textbf{0.000} & \textbf{1.000} & \textbf{0.000} & \textbf{1.000} \\
 & Point-Delta-JEPA & \textbf{0.000} & 0.999 & \textbf{0.000} & \textbf{1.000} \\
\midrule
\multirow{2}{*}{Elbow Position}
 & Point-LeWM & \textbf{0.000} & \textbf{1.000} & \textbf{0.000} & \textbf{1.000} \\
 & Point-Delta-JEPA & \textbf{0.000} & 0.999 & \textbf{0.000} & \textbf{1.000} \\
\midrule
\multirow{2}{*}{Joint Angles}
 & Point-LeWM & \textbf{0.000} & \textbf{1.000} & \textbf{0.000} & \textbf{1.000} \\
 & Point-Delta-JEPA & 0.001 & 0.998 & \textbf{0.000} & \textbf{1.000} \\
\bottomrule
\end{tabular}
\end{table}

% Cube probing
\begin{table}[t]
\centering
\caption{\textbf{Physical latent probing results on OGB-Cube.}
Positions are 3D coordinates in the sensor frame, stacked across the five arm joints for
\emph{Joint Positions}. Yaw is encoded as a heading vector in the sensor frame, and joint velocity is
the raw six-dimensional vector in rad/s. Lower MSE and higher $R^2$ indicate better representation
quality. The near-zero yaw rows reflect label unidentifiability (a cube is
rotationally symmetric, and the wrist returns few points), not representation
failure, as discussed in the text.}
\label{tab:probe-cube}
\begin{tabular}{llcccc}
\toprule
& & \multicolumn{2}{c}{\textbf{Linear}} & \multicolumn{2}{c}{\textbf{MLP}} \\
\cmidrule(lr){3-4} \cmidrule(lr){5-6}
\textbf{Property} & \textbf{Method} & \textbf{MSE} $\downarrow$ & $R^2$ $\uparrow$ & \textbf{MSE} $\downarrow$ & $R^2$ $\uparrow$ \\
\midrule
\multirow{2}{*}{Joint Positions}
 & Point-LeWM & \textbf{0.000} & 0.992 & \textbf{0.000} & 0.999 \\
 & Point-Delta-JEPA & \textbf{0.000} & \textbf{0.998} & \textbf{0.000} & \textbf{1.000} \\
\midrule
\multirow{2}{*}{Joint Velocity}
 & Point-LeWM & 0.142 & 0.021 & 0.130 & 0.102 \\
 & Point-Delta-JEPA & \textbf{0.071} & \textbf{0.507} & \textbf{0.054} & \textbf{0.626} \\
\midrule
\multirow{2}{*}{End-Effector Position}
 & Point-LeWM & \textbf{0.000} & 0.992 & \textbf{0.000} & 0.999 \\
 & Point-Delta-JEPA & \textbf{0.000} & \textbf{0.999} & \textbf{0.000} & \textbf{1.000} \\
\midrule
\multirow{2}{*}{End-Effector Yaw}
 & Point-LeWM & 0.330 & $-0.001$ & 0.332 & $-0.006$ \\
 & Point-Delta-JEPA & \textbf{0.321} & \textbf{0.029} & \textbf{0.321} & \textbf{0.026} \\
\midrule
\multirow{2}{*}{Block Position}
 & Point-LeWM & 0.001 & 0.956 & \textbf{0.000} & 0.996 \\
 & Point-Delta-JEPA & \textbf{0.000} & \textbf{0.996} & \textbf{0.000} & \textbf{0.999} \\
\midrule
\multirow{2}{*}{Block Yaw}
 & Point-LeWM & \textbf{0.333} & $\mathbf{-0.001}$ & \textbf{0.335} & $\mathbf{-0.006}$ \\
 & Point-Delta-JEPA & \textbf{0.333} & $\mathbf{-0.001}$ & \textbf{0.335} & $\mathbf{-0.006}$ \\
\bottomrule
\end{tabular}
\end{table}

\paragraph{Real-world latent probing.}
\label{app:rh20t}
\begin{figure}[!b]
    \centering
    \includegraphics[width=1\linewidth]{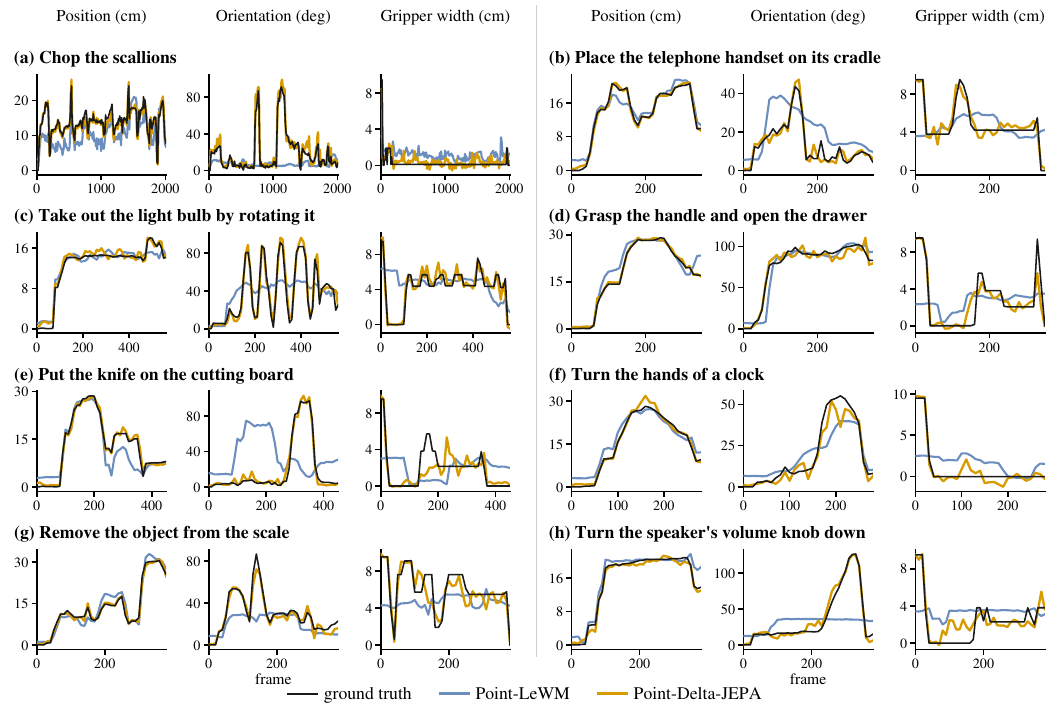}
   \caption{Gripper position, orientation and width predicted by MLP probes on frozen
    latents for eight held-out RH20T episodes from different tasks. Point-Delta-JEPA follows the ground truth closely, while Point-LeWM often misses orientation changes.}
    \label{fig:rh20t_eight_samples}
\end{figure}
So far, all our experiments used simulated data with perfect point clouds, actions, and states. To test whether the latents of our models still encode the task-relevant state on sparse and noisy real point clouds, we ran a probing experiment with Point-LeWM and Point-Delta-JEPA trained entirely on real recordings. We use RH20T \citep{fang2024rh20t}, a real-world manipulation dataset with RGB and depth images, robot states and actions, showing a 7-DoF Flexiv arm interacting with many different objects. We use 1,999 teleoperated episodes (810,348 frames) of configuration 1, recorded by its fixed front-facing RGB-D camera. Each depth image is cleaned before use, which includes removing pixels outside 0.1 to 2.5 m, correcting compression artifacts, and removing flying pixels between foreground and background surfaces. Each cleaned depth frame is then back-projected using the camera intrinsics, and the resulting point cloud is transformed into the robot's base frame. Following \citet{huang2026pointworld}, we crop the points to the robot's workspace. Finally, we randomly subsample each cloud to 10,000 points to match the format of our simulated datasets. Actions are defined as deltas between consecutive frames and follow the OGB-Cube format (dx, dy, dz, dyaw, gripper), extended by droll and dpitch because RH20T's end-effector is not kept gravity-aligned. The states used for probing are the gripper position, orientation, opening width, and the seven joint angles. Orientation is stored in the continuous 6D rotation representation \citep{zhou2019continuity}, which avoids the discontinuities of Euler angles and quaternions. We train Point-LeWM and Point-Delta-JEPA on the resulting dataset with the same architectures and hyperparameters as on the simulated data, for 50 epochs on two A100 GPUs. No real-data-specific changes to the models are needed, since adapting to real data happens entirely in the data pipeline. We then freeze the encoders and train linear and MLP probes on their latents to predict the gripper position, orientation, opening width, and the seven joint angles. Episodes are split 72/8/20 into training, validation, and test sets, so test episodes are never seen during probing.

\begin{table}[h]
\centering
\caption{\textbf{Physical latent probing results on RH20T.}
Positions are 3D coordinates in the robot base frame. Orientation is the full 3D rotation of the
end-effector (6D representation), and joint positions are the seven raw joint angles. Error is
reported in physical units: mean 3D distance for the end-effector position, mean absolute error for
the gripper width and the joint angles, and mean rotation angle between predicted and true
orientation. Lower error and higher $R^2$ indicate better representation quality.}
\label{tab:probe-rh20t}
\begin{tabular}{llcccc}
\toprule
& & \multicolumn{2}{c}{\textbf{Linear}} & \multicolumn{2}{c}{\textbf{MLP}} \\
\cmidrule(lr){3-4} \cmidrule(lr){5-6}
\textbf{Property} & \textbf{Method} & \textbf{Error} $\downarrow$ & $R^2$ $\uparrow$ & \textbf{Error} $\downarrow$ & $R^2$ $\uparrow$ \\
\midrule
\multirow{2}{*}{\shortstack[l]{Joint Positions\\(deg)}}
 & Point-LeWM & 8.49 & 0.367 & 7.77 & 0.435 \\
 & Point-Delta-JEPA & \textbf{4.95} & \textbf{0.767} & \textbf{3.57} & \textbf{0.857} \\
\midrule
\multirow{2}{*}{\shortstack[l]{End-Effector Position\\(cm)}}
 & Point-LeWM & 7.01 & 0.738 & 5.48 & 0.825 \\
 & Point-Delta-JEPA & \textbf{2.87} & \textbf{0.946} & \textbf{2.13} & \textbf{0.964} \\
\midrule
\multirow{2}{*}{\shortstack[l]{End-Effector Orientation\\(deg)}}
 & Point-LeWM & 20.34 & 0.184 & 19.03 & 0.221 \\
 & Point-Delta-JEPA & \textbf{10.80} & \textbf{0.709} & \textbf{7.86} & \textbf{0.828} \\
\midrule
\multirow{2}{*}{\shortstack[l]{Gripper Width\\(cm)}}
 & Point-LeWM & 2.49 & 0.091 & 2.32 & 0.143 \\
 & Point-Delta-JEPA & \textbf{1.48} & \textbf{0.619} & \textbf{1.10} & \textbf{0.728} \\
\bottomrule
\end{tabular}
\end{table}

Table~\ref{tab:probe-rh20t} and Figure~\ref{fig:rh20t} show the results on the 400 held-out test episodes. Point-Delta-JEPA's latents encode the full gripper state: from a single frame, an MLP probe recovers the gripper position to within 2.1 cm ($R^2 = 0.96$), its orientation to within 7.9° ($R^2 = 0.83$), its opening width to within 1.1 cm ($R^2 = 0.73$), and the joint angles to within 3.6° ($R^2 = 0.86$). Even a linear probe recovers most of this information (2.9 cm, 10.8°, 1.5 cm), so it is directly accessible in the latent and not only reachable through a nonlinear decoder. These errors are small compared with the roughly one-metre workspace, although the input clouds are noisy, sparse and unordered, and the scenes and objects change from episode to episode. Point-LeWM also encodes the gripper position (5.5 cm, $R^2 = 0.83$), but hardly the orientation (19.0°, $R^2 = 0.22$) or the opening width (2.3 cm, $R^2 = 0.14$). Over a held-out episode (Figure~\ref{fig:rh20t}a), Point-Delta-JEPA's decoded state follows the true gripper motion closely, while Point-LeWM's stays nearly flat for orientation and width.

\paragraph{Latent surprise}
\label{app:rh20t-surprise}

\begin{figure}[t]
    \centering
    \includegraphics[angle=90, width=1\linewidth]{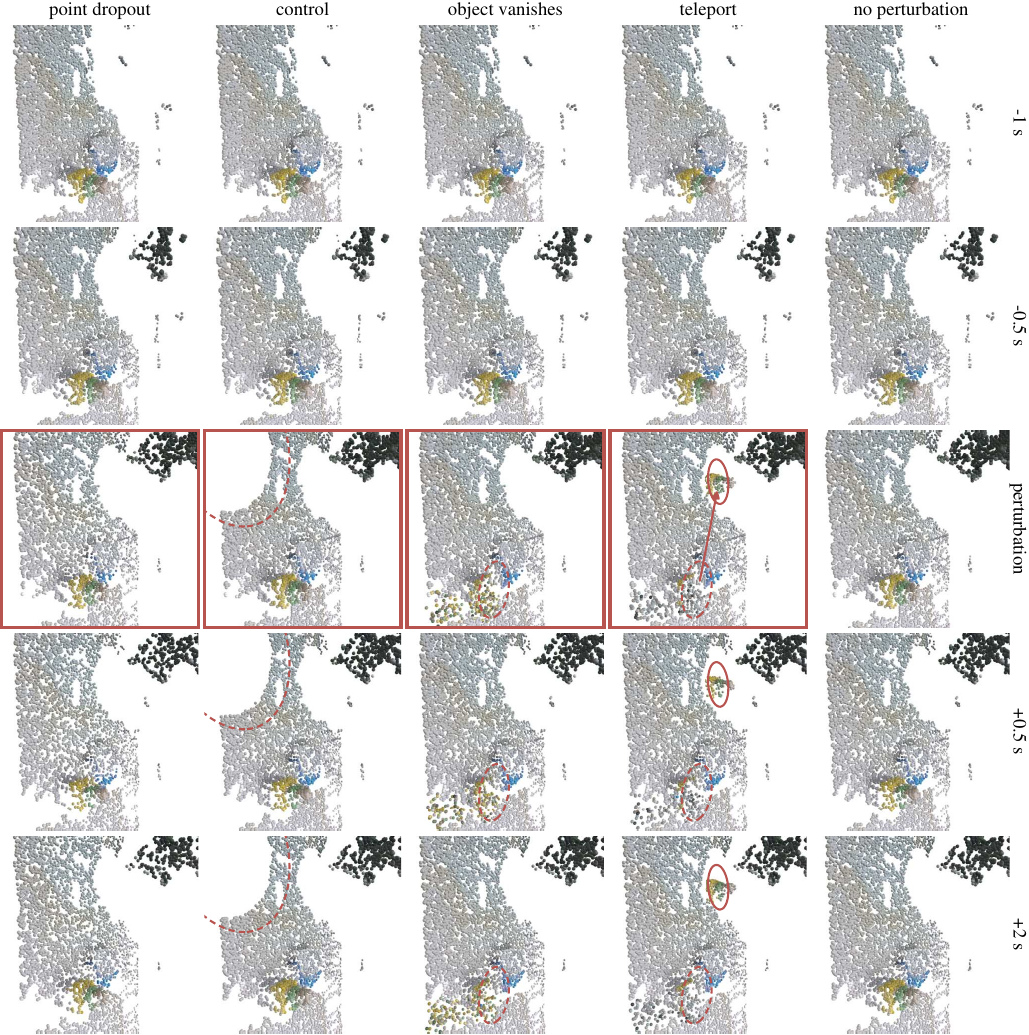}
    \caption{\textbf{Perturbed point clouds.} One RH20T episode before and after the
    perturbation (red frame), shown for each perturbation in Figure~\ref{fig:surprise_rh20t}.
    Circles mark the object's original and new position and the removed patch (control);
    the arrow shows the teleport.}

    \label{fig:surprise_frames_rh20t}
\end{figure}

Following LeWM, we measure surprise as the 1-step latent prediction error and apply
a perturbation at $t=0$ that persists in all later frames. We edit the point clouds
directly: the object the robot later grasps is moved 15--35\,cm on the table
(teleport) or removed (vanish), with its former area filled with synthetic table
points. As controls, we remove the same number of points elsewhere or drop 50\% of
all points. We use the 32 held-out episodes in which the object can be edited cleanly
(Figure~\ref{fig:surprise_frames_rh20t}). Both models react strongly to teleport and
vanish (Wilcoxon signed-rank test, $p<10^{-8}$) and only weakly to the controls
(Figure~\ref{fig:surprise_rh20t}).

\subsection{Point cloud perturbations}
\label{app:degredation}

\paragraph{Protocol.}
Perturbations are applied at evaluation time only, downstream of a faithful
scan. No model ever saw a degraded cloud in training. Test-time-only
degradation is the probe matched to our question: it asks whether latents
trained on clean geometry anchor to structure a worse sensor still measures,
the situation of a model deployed on hardware noisier than its training data.
Degrading the training data instead would measure whether the objectives can
absorb noise, a property of training rather than of the learned
representation. \emph{Noise} is along-ray rather than isotropic, since a
range sensor errs in the range it reports and each return therefore moves
along its own line of sight. Its standard deviation $\sigma$ is set as a
fraction of the per-frame bounding-box diagonal of the valid returns. The relative scale is what makes the ablation comparable
across environments: an absolute $\sigma$ of 2\,cm is a light haze on the
4.9\,m Push-T arena and a catastrophe on the 1.4\,m Reacher workspace.
\emph{Dropout} marks a random fraction of returns as misses, which is what a
lost return physically is, so the dropped points leave through the same
invalid-return filter the pipeline applies to genuine misses. Because the goal
cloud is read from the dataset rather than scanned live, it stays clean under
the first condition (a degraded sensor pursuing a stored target, the
deployment case) and is degraded identically under the second, which tests
the representation itself but no longer distinguishes ``cannot see'' from
``aiming at the wrong thing''. Success rates in
Table~\ref{tab:ablation-compare} are reported relative to each model's own
clean performance, so values may exceed 100 within seed noise.

\begin{table*}[t]
  \centering
  \caption{\textbf{Point cloud degradation.} Planning success rate over 10 seeds with 50 episodes each, reported \emph{relative to each model's own clean performance} (100 = no degradation). The top row gives the absolute clean success rates used as denominators. The two point-cloud world models are shown side by side under each condition, better of the pair in bold. Due to Utonia-WM's long computation times, it is not part of this ablation table. Noise $\sigma$ is a fraction of the frame's scan extent, and dropout is the fraction of returns lost.}
  \resizebox{\textwidth}{!}{%
  \label{tab:ablation-compare}
  \begin{tabular}{lcccccccc}
    \toprule
    & \multicolumn{2}{c}{Two-Room} & \multicolumn{2}{c}{Reacher} & \multicolumn{2}{c}{Push-T} & \multicolumn{2}{c}{OGB-Cube} \\
    \cmidrule(lr){2-3}\cmidrule(lr){4-5}\cmidrule(lr){6-7}\cmidrule(lr){8-9}
    \textbf{Perturbation} & P-LeWM & PD-JEPA & P-LeWM & PD-JEPA & P-LeWM & PD-JEPA & P-LeWM & PD-JEPA \\
    \midrule
    None & 87.0\sd{4.4} & 100.0\sd{0.0} & 80.4\sd{4.6} & 77.6\sd{4.4} & 83.6\sd{3.4} & 70.8\sd{7.7} & 66.0\sd{5.9} & 83.4\sd{3.8} \\
    \midrule
    \multicolumn{9}{l}{\emph{Observation degraded, goal clean}} \\
    \midrule
    Noise $0.25\%$ & \textbf{80.2\sd{9.4}} & 42.8\sd{6.1} & \textbf{96.3\sd{8.2}} & 89.4\sd{6.9} & 1.9\sd{1.9} & \textbf{3.7\sd{2.7}} & \textbf{103.0\sd{10.8}} & 99.8\sd{6.2} \\
    Noise $0.5\%$ & \textbf{52.2\sd{11.5}} & 36.0\sd{6.7} & \textbf{90.8\sd{5.9}} & 62.2\sd{6.1} & 1.2\sd{1.7} & \textbf{4.0\sd{2.0}} & 97.6\sd{11.4} & \textbf{98.3\sd{6.8}} \\
    Dropout $25\%$ & \textbf{99.8\sd{5.4}} & 72.2\sd{6.4} & \textbf{97.5\sd{6.1}} & 90.7\sd{6.5} & 90.7\sd{6.8} & \textbf{94.6\sd{8.8}} & \textbf{101.5\sd{12.7}} & 99.3\sd{4.8} \\
    Dropout $50\%$ & \textbf{98.9\sd{5.1}} & 64.0\sd{6.7} & \textbf{94.8\sd{6.1}} & 87.3\sd{5.6} & 49.0\sd{8.5} & \textbf{52.8\sd{10.5}} & 99.4\sd{12.0} & \textbf{100.7\sd{4.9}} \\
    \midrule
    \multicolumn{9}{l}{\emph{Observation \emph{and} goal degraded}} \\
    \midrule
    Noise $0.25\%$ & \textbf{61.6\sd{5.7}} & 31.0\sd{5.1} & \textbf{91.8\sd{7.8}} & 84.7\sd{6.0} & 3.6\sd{2.9} & \textbf{4.0\sd{3.2}} & \textbf{97.0\sd{12.0}} & 93.5\sd{6.8} \\
    Noise $0.5\%$ & \textbf{37.9\sd{6.0}} & 35.8\sd{6.8} & \textbf{87.6\sd{7.5}} & 58.0\sd{6.7} & \textbf{4.5\sd{2.4}} & 2.5\sd{2.1} & \textbf{89.7\sd{15.9}} & 77.2\sd{9.4} \\
    Dropout $25\%$ & \textbf{96.1\sd{5.9}} & 60.4\sd{6.4} & \textbf{94.8\sd{5.0}} & 89.1\sd{8.7} & \textbf{88.6\sd{6.3}} & 87.0\sd{8.2} & \textbf{99.1\sd{12.3}} & 98.8\sd{4.1} \\
    Dropout $50\%$ & \textbf{92.2\sd{4.6}} & 50.2\sd{5.5} & \textbf{92.3\sd{5.8}} & 75.9\sd{3.2} & 27.4\sd{5.5} & \textbf{35.3\sd{8.6}} & \textbf{101.5\sd{8.5}} & 99.5\sd{4.4} \\
    \bottomrule
  \end{tabular}}
\end{table*}

Beyond the trends discussed in the main text, the table shows that degrading
the goal in addition to the observation costs a further, roughly uniform
margin (e.g.\ Two-Room noise $0.25\%$: 80.2 vs.\ 61.6 relative for \plewm{}),
and that dropout is benign wherever the surviving returns still sample the
entities. Only Push-T, whose pusher is near the ray-budget floor of
Appendix~\ref{app:pc-detail}, breaks under $50\%$ dropout.

\begin{table}[!t]
  \centering
  \caption{\textbf{Effect of the target-to-latent module under joint observation-and-goal degradation.} Planning success rate over 10 seeds with 50 episodes each, reported \emph{relative to each base model's clean performance} (100 = no degradation; clean rates as in Table~\ref{tab:ablation-compare}). \emph{Base} repeats the observation-and-goal-degraded rows of Table~\ref{tab:ablation-compare}; \emph{+MLP} and \emph{+MLP\textsubscript{z}} add the unconditioned and the $\vz_t$-conditioned MLP target encoder of Section~\ref{sec:models} on top of the same backbone, so the goal enters as a clean 3D target instead of a degraded goal cloud. Best of the three variants per model in bold. Noise $\sigma$ is a fraction of the frame's scan extent, and dropout is the fraction of returns lost.}
  \label{tab:ablation-mlp}
  \small
  \setlength{\tabcolsep}{5pt}
  \begin{tabular}{llcccccc}
    \toprule
    & & \multicolumn{3}{c}{P-LeWM} & \multicolumn{3}{c}{PD-JEPA} \\
    \cmidrule(lr){3-5}\cmidrule(lr){6-8}
    \textbf{Env.} & \textbf{Perturbation} & Base & +MLP & +MLP\textsubscript{z} & Base & +MLP & +MLP\textsubscript{z} \\
    \midrule
    \multirow{4}{*}{Two-Room}
      & Noise $0.25\%$ & 61.6\sd{5.7} & 80.0\sd{6.6} & \textbf{80.2\sd{7.0}} & 31.0\sd{5.1} & \textbf{44.4\sd{5.1}} & 41.0\sd{5.6} \\
      & Noise $0.5\%$  & 37.9\sd{6.0} & \textbf{53.8\sd{10.1}} & 53.6\sd{8.9} & 35.8\sd{6.8} & 35.0\sd{5.8} & \textbf{36.4\sd{4.3}} \\
      & Dropout $25\%$ & 96.1\sd{5.9} & \textbf{100.7\sd{5.3}} & 99.3\sd{5.4} & 60.4\sd{6.4} & 71.4\sd{7.5} & \textbf{72.0\sd{6.8}} \\
      & Dropout $50\%$ & 92.2\sd{4.6} & \textbf{99.5\sd{4.5}} & 98.4\sd{5.7} & 50.2\sd{5.5} & \textbf{63.4\sd{3.9}} & 63.0\sd{4.2} \\
    \midrule
    \multirow{4}{*}{Reacher}
      & Noise $0.25\%$ & \textbf{91.8\sd{7.8}} & 90.8\sd{7.1} & 91.0\sd{8.2} & 84.7\sd{6.0} & 86.9\sd{7.6} & \textbf{89.2\sd{10.1}} \\
      & Noise $0.5\%$  & 87.6\sd{7.5} & 86.6\sd{6.5} & \textbf{89.1\sd{5.8}} & \textbf{58.0\sd{6.7}} & 54.6\sd{12.8} & 56.2\sd{12.9} \\
      & Dropout $25\%$ & 94.8\sd{5.0} & \textbf{96.5\sd{9.2}} & \textbf{96.5\sd{8.0}} & 89.1\sd{8.7} & \textbf{96.9\sd{8.9}} & 94.1\sd{7.7} \\
      & Dropout $50\%$ & 92.3\sd{5.8} & \textbf{96.5\sd{8.0}} & 93.8\sd{6.0} & 75.9\sd{3.2} & \textbf{86.9\sd{8.5}} & 83.0\sd{12.6} \\
    \midrule
    \multirow{4}{*}{Push-T}
      & Noise $0.25\%$ & \textbf{3.6\sd{2.9}} & 2.2\sd{2.4} & 2.6\sd{2.9} & \textbf{4.0\sd{3.2}} & \textbf{4.0\sd{3.5}} & 3.1\sd{2.5} \\
      & Noise $0.5\%$  & \textbf{4.5\sd{2.4}} & 2.2\sd{2.4} & 2.2\sd{2.4} & 2.5\sd{2.1} & 3.1\sd{3.7} & \textbf{3.4\sd{3.7}} \\
      & Dropout $25\%$ & 88.6\sd{6.3} & \textbf{91.1\sd{6.0}} & 90.2\sd{8.6} & 87.0\sd{8.2} & \textbf{98.6\sd{7.3}} & 95.2\sd{7.6} \\
      & Dropout $50\%$ & 27.4\sd{5.5} & 49.0\sd{9.0} & \textbf{49.8\sd{6.7}} & 35.3\sd{8.6} & \textbf{51.4\sd{9.0}} & 50.6\sd{7.9} \\
    \midrule
    \multirow{4}{*}{OGB-Cube}
      & Noise $0.25\%$ & 97.0\sd{12.0} & 100.0\sd{11.8} & \textbf{100.6\sd{10.5}} & 93.5\sd{6.8} & 84.9\sd{7.8} & \textbf{98.3\sd{7.6}} \\
      & Noise $0.5\%$  & 89.7\sd{15.9} & \textbf{99.7\sd{11.7}} & 97.6\sd{11.2} & 77.2\sd{9.4} & 79.6\sd{9.5} & \textbf{90.9\sd{8.3}} \\
      & Dropout $25\%$ & 99.1\sd{12.3} & \textbf{100.6\sd{12.0}} & 100.0\sd{10.0} & \textbf{98.8\sd{4.1}} & 84.9\sd{7.4} & 98.6\sd{7.4} \\
      & Dropout $50\%$ & \textbf{101.5\sd{8.5}} & 101.2\sd{12.3} & 100.0\sd{13.0} & 99.5\sd{4.4} & 85.4\sd{5.6} & \textbf{101.0\sd{6.6}} \\
    \bottomrule
  \end{tabular}
\end{table}

\paragraph{Target-to-latent module under degradation.}
Table~\ref{tab:ablation-mlp} extends the joint-degradation rows of
Table~\ref{tab:ablation-compare} to planners that receive their goal through
the target-to-latent module of Section~\ref{sec:models}, with the
unconditioned MLP (+MLP) and the $\vz_t$-conditioned MLP
(+MLP\textsubscript{z}) as target encoders. The base planner encodes the
degraded goal cloud. The module instead receives the clean 3D target and, in
the conditioned case, the latent of the degraded current observation, which
is the deployment case of a noisy sensor pursuing a target given in
coordinates. All numbers remain relative to the base model's clean success
rate, so the clean cost of the module itself (Table~\ref{tab:target2latent})
is included in every cell.

Three regularities emerge. First, the conditioned module is at or above the
base model in 19 of 24 cells, and the five exceptions lie within seed noise,
three of them at the Push-T noise floor and two on OGB-Cube at most 1.5
points below. The largest gains appear where the base model loses most:
Two-Room under range noise (+19 and +16 points for \plewm{}, +10 for
\pdjepa{} at $0.25\%$), Two-Room under dropout for \pdjepa{} (+12 and +13),
Push-T under $50\%$ dropout (+22 and +15) and OGB-Cube under $0.5\%$ noise
(+8 and +14). Second, the module recovers the observation-only level. In all
but one cell, the conditioned module lands within seed noise of the
corresponding \emph{observation degraded, goal clean} row of
Table~\ref{tab:ablation-compare}, the exception being \pdjepa{} on OGB-Cube
at $0.5\%$ noise (90.9 vs.\ 98.3). Specifying the goal as a low-dimensional
target thus removes the sensor from the goal channel, and what remains is the
robustness of the observation encoder alone. Consequently, the module does not
repair failures on the observation side. Range noise on Push-T erases the
thin task entities from the scan (Section~\ref{sec:experiments}) and stays
below 5 relative points for every variant. Third, conditioning on $\vz_t$
matters under degradation exactly where it matters on clean data. On Two-Room
and Push-T the unconditioned and conditioned MLPs are tied within seed noise.
On OGB-Cube the unconditioned MLP sits at 80 to 85 relative points for
\pdjepa{} regardless of the perturbation, below the base in three of four
cells. This offset is not a degradation effect. It matches the unconditioned
MLP's clean deficit on that environment (72.0 vs.\ 84.2 absolute in
Table~\ref{tab:target2latent}, about 86 relative), which arises because the
OGB-Cube target leaves the arm configuration unconstrained. Taken together,
planning toward a learned target latent is at least as robust to sensor
degradation as planning toward an encoded goal cloud, and substantially more
robust whenever the goal cloud itself is degraded.

\subsubsection{Commanding goals the expert never aimed at}
\label{sec:cube_grid}

\begin{figure}[t]
\centering
\begin{subfigure}{.5\textwidth}
  \centering
  \includegraphics[width=\linewidth]{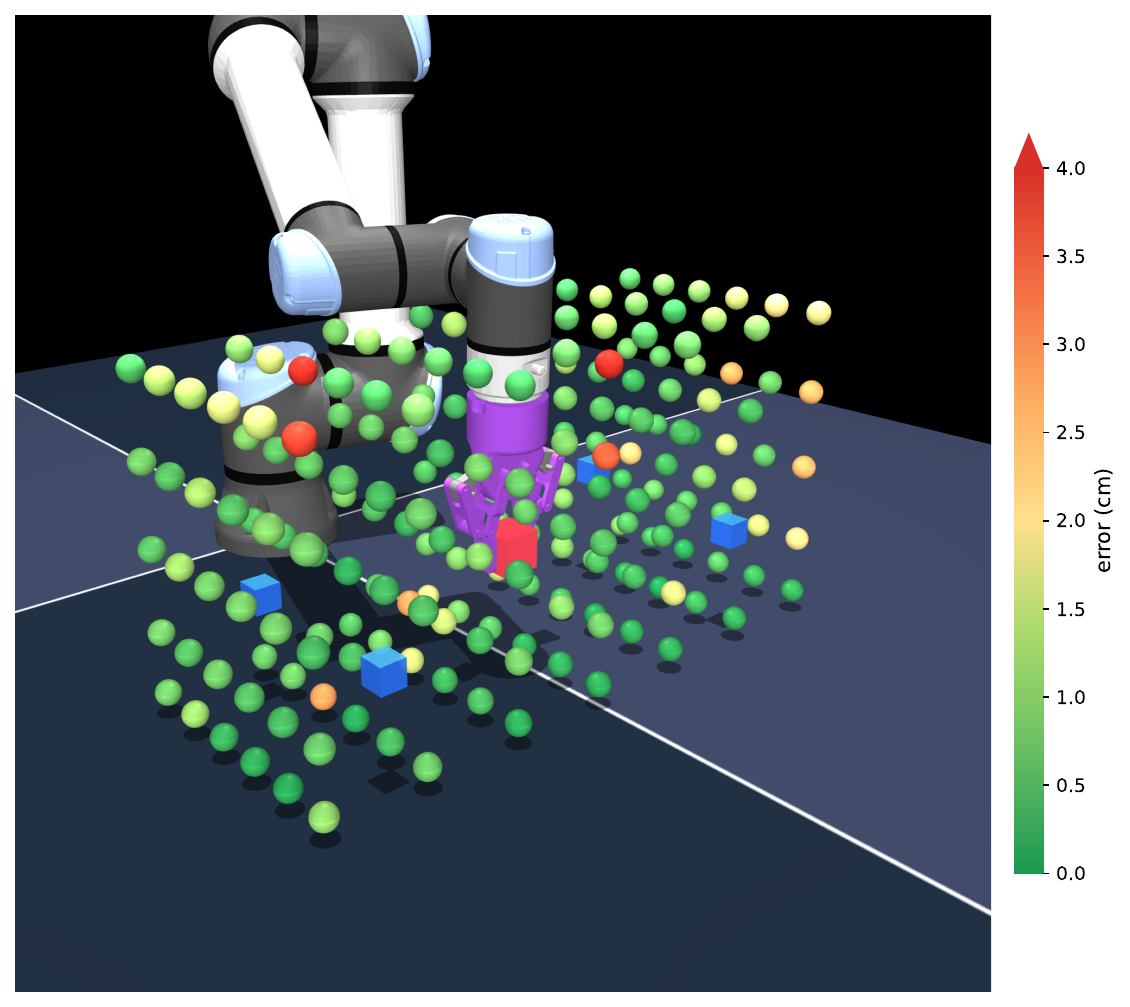}
  \caption{Rebuilt goal cloud: 0.66\,cm median.}
  \label{fig:cube_grid_gc}
\end{subfigure}%
\begin{subfigure}{.5\textwidth}
  \centering
  \includegraphics[width=\linewidth]{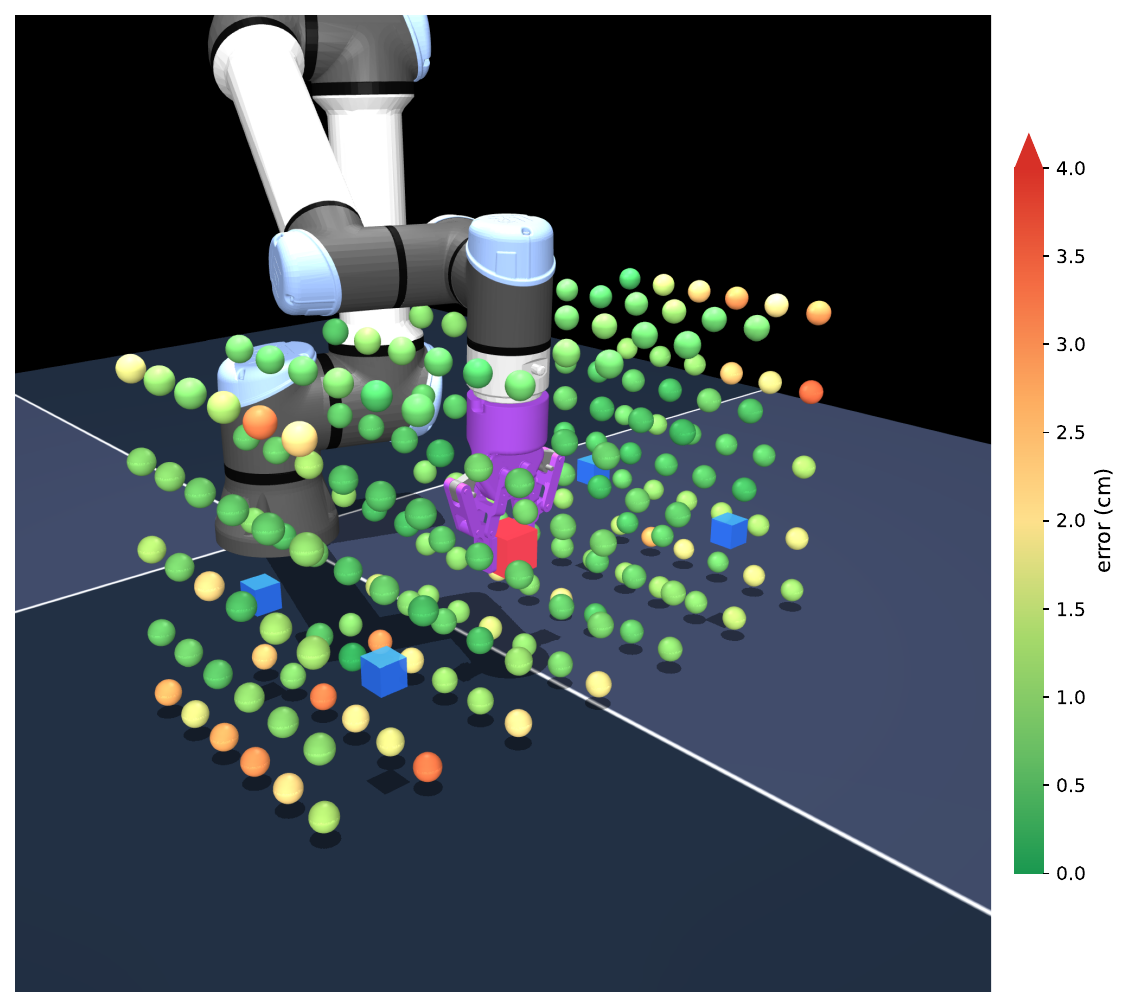}
  \caption{Target-to-latent: 0.77\,cm median.}
  \label{fig:cube_grid_t2l}
\end{subfigure}
\caption{\textbf{Specified cube goals.} 201 cube positions at five heights ($z = 2, 8, 16, 24,
32$\,cm), each approached from five grasped start poses with the frozen
Point-Delta-JEPA point-cloud world model. Target-to-latent maps the
position, together with the latent of the current observation, to the goal
latent. Planner and goal latents operate on LiDAR point clouds; the rendered
scene is shown only for legibility. Colour is the median over the five starts
of the smallest cube-to-goal distance, from 0\,cm (green) to the 4\,cm success
threshold (red). Blue cubes mark the four corner start poses, and the red cube
in the gripper marks the center start. Only $z = 2$\,cm occurs as a goal in
training.}
\label{fig:cube_grid}
\end{figure}

Every cube goal in the dataset is a placement on the table, so a cube held
above it is a goal the expert never aimed at. We command 201 positions on five
height layers up to 32\,cm, each from five grasped start poses (blue in
Figure~\ref{fig:cube_grid}), and compare against the strongest goal-cloud
baseline: the specified state rendered and re-scanned in simulation. On all
never-seen heights, target-to-latent matches it (0.68 vs.\ 0.67\,cm,
$p = 0.94$, Table~\ref{tab:cube_grid}) from the coordinates alone.
Each method has one weak region: target-to-latent on the table layer, where
the two goals differ (held vs.\ released cube), and the goal cloud in the far
corners at height, which lie outside the LiDAR's field of view.

\begin{table}[t]
\centering
\caption{\textbf{Accuracy on unseen goals.} Cube goals at heights the expert never specified. Each 3-D
position is approached from five grasped start poses with the same
Point-Delta-JEPA point-cloud world model and CEM planner, so only the goal
latent differs between target-to-latent, which predicts it from the
position and the current latent, and the rebuilt goal cloud, which encodes the state rendered and re-scanned in simulation. Error is the cube's
closest approach to the goal, and success uses the environment's 4\,cm
threshold. $\Delta$ gives the median paired difference between the two methods
with a bootstrap 95\,\% interval, where negative values favour target-to-latent.
Over all elevated goals the two are indistinguishable (Wilcoxon signed-rank,
$p = 0.94$).}
\label{tab:cube_grid}
\setlength{\tabcolsep}{4.5pt}
\begin{tabular}{lrcccrr}
\toprule
 & & \multicolumn{2}{c}{Median error (cm)} & & \multicolumn{2}{c}{Success (\%)} \\
\cmidrule(lr){3-4}\cmidrule(lr){6-7}
Goal height & $n$ & T2L & Goal cloud & $\Delta$ [95\,\% CI] & T2L & Goal cloud \\
\midrule
8\,cm  & 165 & 0.76 & 0.68 & $+0.05$ [$-0.06$, $+0.14$] & 96 & 95 \\
16\,cm & 210 & 0.58 & 0.51 & $+0.12$ [$+0.03$, $+0.22$] & 97 & 95 \\
24\,cm & 210 & 0.63 & 0.68 & $-0.07$ [$-0.15$, $+0.01$] & 98 & 94 \\
32\,cm & 210 & 0.85 & 0.85 & $-0.02$ [$-0.18$, $+0.09$] & 96 & 91 \\
\cmidrule(lr){1-7}
All & 795 & 0.68 & 0.67 & $+0.02$ [$-0.05$, $+0.07$] & 97 & 94 \\
\quad outside sensor view$^{\dagger}$ & 35 & 2.20 & 1.98 & $+0.17$ [$-0.40$, $+0.73$] & 83 & 89 \\
\midrule
Goal input & & \multicolumn{2}{c}{\small pose \quad\;\; point cloud} & & & \\
\bottomrule
\end{tabular}
\vspace{2pt}
\footnotesize
$^{\dagger}$ Positions beyond the LiDAR's field of view; the rebuilt
goal cloud does not contain the cube.
\end{table}

\subsection{Attention}
\label{app:attention}

\begin{figure}
    \centering
    \includegraphics[width=1\linewidth]{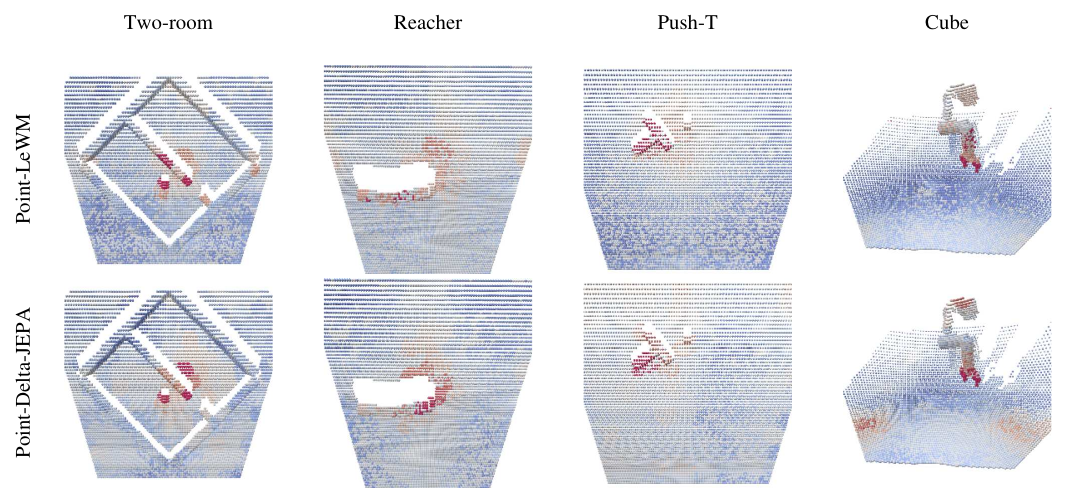}
    \caption{\textbf{Encoder attention.} \textsc{cls}-token attention rollout of the
\plewm{}/\pdjepa{} encoder mapped onto the input points. Warmer colours indicate higher
attention, and points no group token covers are omitted. Columns are
environments, rows the two models. Attention is normalised per point cloud.
Utonia-WM is absent because its frozen PTv3 encoder has no \textsc{cls} token
and attends only locally within serialised patches, so it admits no comparable
attention map.}
    \label{fig:attention}
\end{figure}

Figure~\ref{fig:attention} visualizes where the two end-to-end encoders look:
the \textsc{cls}-token attention is propagated through the ViT trunk by
attention rollout, assigned from each group token to the points it covers, and
normalised per cloud. Both models place their
attention mass on the moving entities. In all four environments \plewm{} is
sharply peaked on the entities, while \pdjepa{} allots visibly more
attention to ground and table.

\subsection{Multi-Modal Planning}
\label{app:multi-modal-planning}

Table \ref{tab:fusion} shows the numerical results of the multi-modal planning successes discussed in Section \ref{sec:multi-modal-planning}.

\begin{table}[h]
  \centering
  \caption{Cost-level fusion of the point cloud and RGB world models at planning time. Planning success rate (\%) as a function of the RGB share $w$ in the candidate cost $c = (1-w)\,\tilde c_{\text{points}} + w\,\tilde c_{\text{RGB}}$, where $\tilde c$ denotes the per-iteration z-scored cost. $w=0$ is the point-cloud model alone and $w=1$ the RGB model alone. Mean $\pm$ standard deviation over ten evaluation seeds with 50 episodes each. The best value per row is in bold.}
  \label{tab:fusion}
  \setlength{\tabcolsep}{3.5pt}
  \small
  \begin{tabular}{@{}llccccccc@{}}
    \toprule
    & & \multicolumn{7}{c}{RGB share $w$} \\
    \cmidrule(l){3-9}
    Env. & Model & 0 (points) & 0.1 & 0.25 & 0.5 & 0.75 & 0.9 & 1 (RGB) \\
    \midrule
    \multirow{2}{*}{OGB-Cube}
      & P-LeWM  & 64.0\sd{7.2} & 72.2\sd{6.1} & 73.0\sd{7.6} & \textbf{74.4}\sd{5.6} & 72.4\sd{7.8} & 71.2\sd{8.7} & 68.8\sd{9.2} \\
      & PD-JEPA & 83.0\sd{3.7} & 84.4\sd{3.0} & 84.4\sd{4.4} & \textbf{86.2}\sd{4.3} & 82.6\sd{4.4} & 84.4\sd{2.8} & 83.4\sd{3.1} \\
    \midrule
    \multirow{2}{*}{Two-Room}
      & P-LeWM  & \textbf{87.8}\sd{4.4} & 87.0\sd{4.6} & 86.6\sd{4.1} & 85.4\sd{4.4} & 84.4\sd{5.0} & 83.8\sd{5.5} & 83.6\sd{5.9} \\
      & PD-JEPA & \textbf{100.0}\sd{0.0} & \textbf{100.0}\sd{0.0} & \textbf{100.0}\sd{0.0} & \textbf{100.0}\sd{0.0} & \textbf{100.0}\sd{0.0} & 99.4\sd{1.3} & 99.2\sd{1.4} \\
    \midrule
    \multirow{2}{*}{Push-T}
      & P-LeWM  & 83.8\sd{2.0} & 88.2\sd{4.8} & 89.0\sd{3.6} & \textbf{89.6}\sd{4.2} & 88.2\sd{3.5} & 88.8\sd{3.2} & 86.4\sd{4.0} \\
      & PD-JEPA & 70.0\sd{6.9} & 81.6\sd{5.3} & 87.0\sd{4.3} & 94.4\sd{3.0} & \textbf{95.6}\sd{2.5} & 94.4\sd{2.8} & 94.2\sd{4.7} \\
    \midrule
    \multirow{2}{*}{Reacher}
      & P-LeWM  & 77.8\sd{6.1} & 79.4\sd{4.8} & 83.2\sd{5.1} & 83.4\sd{3.7} & \textbf{83.6}\sd{5.9} & 80.8\sd{4.3} & 82.0\sd{7.8} \\
      & PD-JEPA & 74.0\sd{4.3} & 79.2\sd{7.7} & 81.8\sd{5.0} & 84.6\sd{4.6} & 80.4\sd{7.2} & 83.8\sd{5.2} & \textbf{85.0}\sd{1.9} \\
    \bottomrule
  \end{tabular}
\end{table}

\section{Statistical tests}
\label{app:stats}

The tests compare the models within each family of Table~\ref{tab:main},
the image model against its point-cloud counterpart. The frozen-encoder
family adds a second comparison, \utoniawm{} against \voxwm{}, which share
voxel grid, predictor and projection and differ only in whether the voxel
features are pretrained or training-free. Every model is scored on the same
ten evaluation seeds, and each seed fixes the same 50 start--goal windows
for every model, so the tests are paired by seed ($n=10$). Pairing removes
the seed-to-seed variation in episode difficulty from the comparison. Each
model is a single trained checkpoint, so the tests capture the variance of
evaluation over episodes, not the variance across training runs.

We follow the standard protocol for comparing two methods over several
tasks \citep{demsar2006statistical}. The paired $t$-test on the ten per-seed
differences is the primary test, reported with the 95\% confidence interval
of the mean difference. The Wilcoxon signed-rank test
\citep{wilcoxon1945individual} repeats the comparison without assuming
normally distributed differences, which ten seeds cannot verify. Each
comparison spans four benchmarks, so all $p$-values are Holm--Bonferroni
corrected over them \citep{holm1979simple}. The correction keeps the
probability of any false positive within a comparison at $0.05$ and is
uniformly more powerful than the plain Bonferroni correction. A
non-significant difference is not evidence of parity, so wherever the text
states that two models do not differ, the confidence interval bounds how
large their difference can be.

\begin{table}[t]
\centering
\caption{\textbf{Paired tests within the families of Table~\ref{tab:main}.}
Difference of means $\Delta$ (points, first model minus second) with its
95\% confidence interval, and Holm--Bonferroni corrected $p$-values of the
paired $t$-test and the Wilcoxon signed-rank test over the four benchmarks
of each comparison ($n=10$ seeds). With ten seeds, the exact Wilcoxon $p$
cannot fall below $0.002$, or $0.008$ after correction.}
\label{tab:stats}
\begin{tabular}{@{}lcccc@{}}
\toprule
& $\Delta$ & 95\% CI & $t$-test $p$ & Wilcoxon $p$ \\
\midrule
\multicolumn{5}{@{}l}{\emph{SIGReg:} \plewm{} $-$ LeWM} \\
Two-Room & $+1.4$ & $[-2.0, +4.8]$ & $0.75$ & $0.86$ \\
Reacher  & $-1.6$ & $[-5.9, +2.7]$ & $0.75$ & $0.86$ \\
Push-T   & $-2.8$ & $[-5.2, -0.4]$ & $0.099$ & $0.11$ \\
OGB-Cube & $-3.8$ & $[-8.1, +0.5]$ & $0.24$ & $0.30$ \\
\midrule
\multicolumn{5}{@{}l}{\emph{Action reconstruction:} \pdjepa{} $-$ Delta-JEPA} \\
Two-Room & $+0.2$ & $[-0.3, +0.7]$ & $0.34$ & $1.00$ \\
Reacher  & $-8.4$ & $[-10.4, -6.4]$ & $<0.001$ & $0.008$ \\
Push-T   & $-23.4$ & $[-29.4, -17.4]$ & $<0.001$ & $0.008$ \\
OGB-Cube & $+3.2$ & $[-1.3, +7.7]$ & $0.29$ & $0.50$ \\
\midrule
\multicolumn{5}{@{}l}{\emph{Frozen encoder:} \utoniawm{} $-$ DINO-WM} \\
Two-Room & $-5.8$ & $[-7.2, -4.4]$ & $<0.001$ & $0.008$ \\
Reacher  & $-5.8$ & $[-11.4, -0.2]$ & $0.044$ & $0.047$ \\
Push-T   & $-29.4$ & $[-37.0, -21.8]$ & $<0.001$ & $0.008$ \\
OGB-Cube & $-7.4$ & $[-11.4, -3.4]$ & $0.005$ & $0.012$ \\
\midrule
\multicolumn{5}{@{}l}{\emph{Voxel features:} \utoniawm{} $-$ \voxwm{}} \\
Two-Room & $+10.2$ & $[+6.9, +13.5]$ & $<0.001$ & $0.008$ \\
Reacher  & $-0.6$ & $[-5.1, +3.9]$ & $0.77$ & $0.74$ \\
Push-T   & $-9.2$ & $[-12.6, -5.8]$ & $<0.001$ & $0.008$ \\
OGB-Cube & $+6.0$ & $[+2.3, +9.7]$ & $0.010$ & $0.031$ \\
\bottomrule
\end{tabular}
\end{table}

\paragraph{SIGReg.} No benchmark separates \plewm{} from LeWM after
correction under either test. The 2.8-point Push-T deficit is significant
only before correction ($t$-test $p=0.025$, Wilcoxon $p=0.027$). The 95\%
intervals bound the difference to $[-8.1,+4.8]$ points over all four
benchmarks and to within $\pm6$ points on Two-Room, Reacher and Push-T.

\paragraph{Action reconstruction.} \pdjepa{} trails Delta-JEPA on Reacher
($-8.4$) and Push-T ($-23.4$), significant under both tests. It does not
differ from it on Two-Room and OGB-Cube, with intervals $[-0.3,+0.7]$ and
$[-1.3,+7.7]$. On Two-Room the two models differ on a single seed, which
leaves the signed-rank test without power, while the interval is narrow.

\paragraph{Frozen encoder.} \utoniawm{} trails DINO-WM on all four
benchmarks, significant after correction under both tests, with Reacher
borderline ($t$-test $p=0.044$, Wilcoxon $p=0.047$). Against \voxwm{},
\utoniawm{} is better on Two-Room ($+10.2$) and OGB-Cube ($+6.0$) and worse
on Push-T ($-9.2$), all significant under both tests, and does not differ
on Reacher ($-0.6$, interval $[-5.1,+3.9]$).

\paragraph{Correction scope.} We correct within each comparison, since each
answers its own question. Correcting over all 16 tests instead changes
three decisions: the frozen-encoder gap on Reacher loses significance under
both tests, and the OGB-Cube gaps of the frozen-encoder and voxel-feature
comparisons lose it under the Wilcoxon test.

\section{Broader impacts}
\label{app:impact}

This work is foundational research on world-model learning in simulated
tabletop environments and is not tied to a particular application or
deployment. Its societal effects are those generic to improved robot autonomy.
Range-based world models could benefit assistive and industrial robots that
must operate where cameras fail, and lower the cost of adapting robots to new
tasks. Conversely, more capable autonomous systems can displace labor and
could be misused, for example in surveillance platforms that favor range
sensors precisely because they work in darkness. The models released here are
trained on four simulated benchmarks and pose no direct risk, and we see no
mitigations required beyond the norms of open release.

\end{document}